\documentclass[journal]{IEEEtran}

\usepackage{color, array, amsthm}
\usepackage{graphicx}
\usepackage{subcaption}
\usepackage{amssymb}
\usepackage{xcolor}

\usepackage{multirow}

\usepackage{algorithm}
\usepackage{algpseudocode}
\usepackage{amsmath}
\usepackage{hyperref}
\hypersetup{
    pdfborder={0 0 0}  
}
\usepackage{makecell}

\usepackage{adjustbox} 
\usepackage{tikz}
\usepackage{mwe} 
\usetikzlibrary{shapes, arrows, positioning, shapes.geometric, fit, arrows.meta, calc, backgrounds}
\tikzset{
  img/.style   ={draw,rounded corners=2pt,minimum width=3.6cm,minimum height=2.2cm,inner sep=0pt},
  module/.style={draw,rounded corners=3pt,minimum height=1cm,minimum width=1.5cm,align=center,font=\small},
  fe/.style={
    draw=black,
    thick,
    fill=green!20,
    align=center,
    inner sep=2pt,
    minimum width=1.2cm, 
    minimum height=1.8cm,
    shape=trapezium,
    trapezium left angle=70,
    trapezium right angle=70,
    trapezium stretches=true
  },
  stff/.style  ={module,fill=blue!15},
  sppf/.style  ={module,fill=violet!15},
  c2psa/.style ={module,fill=violet!25},
  neck/.style  ={module,fill=yellow!25},
  head/.style  ={module,fill=yellow!35},
  flow/.style  ={-Latex,thick},
  bigbox/.style={draw,rounded corners=12pt,fill=orange!8,thick,inner sep=12pt}
}

\begin{document}

\title{MF-UAVPose6D: A Model-Free Monocular 6-DoF Pose Estimation Framework for Fixed-Wing UAVs} 

\author{Juanqin Liu, Leonardo Plotegher, Eloy Roura and Shaoming He\textsuperscript{*}
\thanks{This work was supported by the Technology Innovation Institute under Contract No. TII/ARRC/2154/2023.}
\thanks{Juanqin Liu and Shaoming~He are with the School of Aerospace Engineering, Beijing Institute of Technology, Beijing 100081, China.}
\thanks{Leonardo Plotegher and Eloy Roura are with the Autonomous Robotics Research Centre, Technology Innovation Institute, P.O.Box: 9639, Masdar City, Abu Dhabi, United Arab Emirates.}
\thanks{\textsuperscript{*}Corresponding Author. Email: \texttt{shaoming.he@bit.edu.cn}.}
}

\maketitle

\begin{abstract}
For uncrewed aerial vehicles (UAVs), estimating six-degree-of-freedom (6-DoF) poses is essential for airspace situational awareness, target tracking, and counter-UAV operations. However, non-cooperative targets usually lack computer-aided design (CAD) models and keypoint priors, making existing model-based or keypoint-matching methods difficult to apply reliably. To address these challenges, this paper proposes MF-UAVPose6D, a model-free monocular 6-DoF pose estimation framework for fixed-wing UAVs. During inference, the method takes only a single red–green–blue (RGB) image and camera intrinsics as input. It first obtains a stable target anchor through heatmap-guided center localization, introduces a Perspective-Aware Module (PAM) to model observation-ray priors, exploits Dynamic Topological Sampling (DTS) to complement weak structural cues from the wings, fuselage, and tail, and adopts a decoupled translation–rotation pose decoding mechanism to estimate the 6-DoF pose. In addition, we construct the FW-UAV6DPose synthetic dataset, which covers fixed-wing UAV observations across diverse distances, viewpoints, and poses. Experimental results show that MF-UAVPose6D achieves accurate and efficient monocular 6-DoF pose estimation without requiring CAD models, and demonstrates strong robustness in long-range rotation estimation, depth recovery, and joint pose evaluation.

\end{abstract}


\section{Introduction}
For UAVs, 6-DoF pose estimation provides accurate target position and orientation information in 3D space, making it a fundamental perception capability for complex airspace situational awareness and counter-UAV systems. Accurate pose information not only enhances the reliability of target early warning and tracking but also supports heading and maneuver-intent analysis, threat-level assessment, and the formulation of subsequent interception, mitigation, and airspace management strategies \cite{chen2023oblique,kim2021vision,luo2023uav,herrera2025deep,hwang2025dronekey}. However, 6-DoF pose estimation for long-range non-cooperative UAVs remains highly challenging: the absence of prior knowledge regarding target category, scale, and 3D structure makes it difficult to establish robust 2D-3D correspondences. Meanwhile, under monocular imaging, pose, depth, and translation are strongly coupled, further limiting the model’s ability to accurately disentangle spatial position from attitude variations.

\begin{figure}[htbp]
	\centering
	\includegraphics[width=\linewidth]{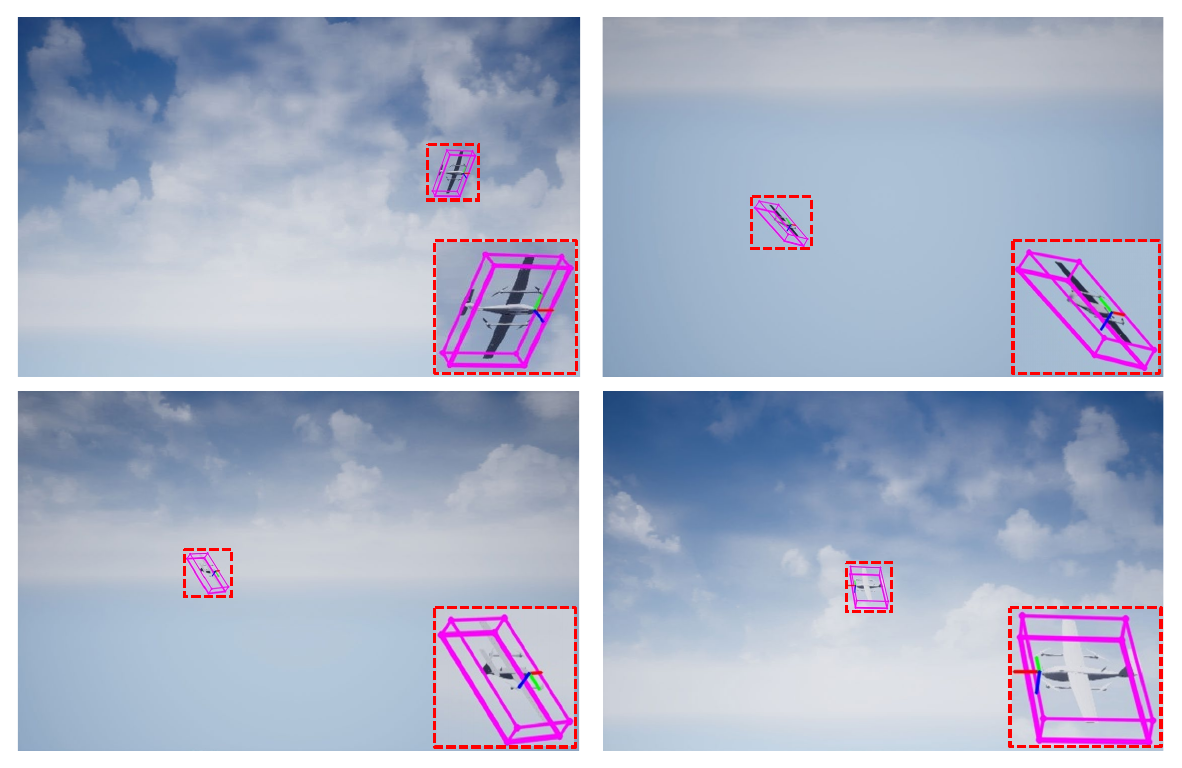}	
	\caption{Examples of Monocular UAV 6-DoF pose Estimation. Our method efficiently infers the 6-DoF pose of UAV targets, including 3D translation and rotation, using only a single RGB image and camera intrinsics.}
	\label{Fig1}	
\end{figure}

Existing pose estimation methods for aerial targets typically rely on known aircraft types, CAD models, structural dimensions, or manually defined keypoints, and recover the pose using geometric solvers such as the Perspective-n-Point (PnP) algorithm \cite{albanis2020dronepose,jeong2025enhanced,pessanha2024fixed}. These methods perform well in close-range scenarios where the target model is available, and keypoints are clearly visible. However, they are difficult to generalize to non-cooperative UAVs with unknown categories and unavailable geometric priors. Long-range imaging further weakens texture and keypoint discriminability, leading to unstable localization, increased rotation errors, and degraded depth estimation. Therefore, how to achieve long-range UAV 6-DoF pose estimation from a single RGB image without relying on target CAD models or 3D keypoint priors remains a valuable yet challenging research problem. As shown in Fig.~\ref{Fig1}, this paper focuses on this model-free monocular pose estimation task, aiming to recover the 3D position and spatial orientation of the target in the camera coordinate system using only a single RGB image and camera intrinsics.

Recent years have witnessed significant progress in model-free and cross-category generalizable 6D object pose estimation. Existing studies have progressively reduced the reliance on instance-level CAD models and precise geometric priors through normalized object coordinates, semantic keypoints, implicit shape representations, cross-view feature alignment, and feature enhancement based on foundation vision models \cite{peng2019pvnet,song2020hybridpose,wang2019normalized,lin2023vi,peng2022self,liu2022gen6d,he2022onepose++}. Furthermore, representative methods such as PoET \cite{jantos2023poet}, MegaPose \cite{labbe2022megapose}, FoundationPose \cite{wen2024foundationpose}, SAM-6D \cite{lin2024sam}, and Any6D \cite{lee2025any6d} indicate that pose estimation for unseen objects is gradually shifting from traditional instance-level geometric matching toward a unified modeling paradigm centered on object-centric representations, coarse-to-fine reasoning, and foundation visual features. However, most of these methods are evaluated on scenarios involving tabletop objects, mechanical parts, indoor robotic grasping, or close-range targets, where objects typically have relatively large image scales and clear appearance structures. In contrast, long-range fixed-wing UAVs exhibit more extreme small-object characteristics, and their pose information mainly relies on weak structural cues such as the slender fuselage, wing orientation, and tail contours \cite{liu2026gl,fang2025ahigh}. These cues are highly susceptible to background noise, scale compression, and boundary blurring in long-range imaging and low-resolution feature maps, making it difficult to directly transfer general close-range object pose estimation methods to long-range UAV 6-DoF pose estimation.

Beyond model design, the scarcity of datasets is another critical factor limiting the development of UAV 6-DoF pose estimation. The collection of real UAV data is often constrained by airspace safety, flight control, sensor synchronization, and environmental disturbances, making it difficult to achieve large-scale coverage of diverse distances, viewpoints, and pose distributions \cite{albanis2020dronepose,liu2025mmfw,yao2024uemm}. In particular, obtaining accurate 6D ground-truth annotations, including 3D position, rotation, camera intrinsics, and projection relationships, is especially challenging in long-range imaging scenarios. Therefore, constructing a synthetic dataset with accurate 6D annotations, controllable distance ranges, and diverse pose distributions is crucial for advancing research on long-range UAV pose estimation.

To address these challenges, this paper proposes MF-UAVPose6D, an end-to-end pose estimation without relying on target CAD models or explicit 3D keypoint matching. The main contributions of this work are as follows:

\begin{itemize}
\item \textbf{Model-free monocular pose estimation framework:} We proposed the MF-UAVPose6D, which estimates the 6-DoF pose of fixed-wing UAVs using only a single RGB image and camera intrinsics, without requiring target CAD models or explicit 3D keypoint matching. This makes the framework more suitable for non-cooperative long-range UAV observation scenarios with limited prior knowledge.

\item \textbf{Geometry-aware end-to-end inference pipeline:} The proposed method first obtains spatial priors of the target through heatmap-guided target center localization, introduces a perspective-aware module to model variations in the line-of-sight angle, and employs dynamic topology sampling to complement structural cues from the wings, fuselage, and tail. Furthermore, a pose decoding mechanism that decouples translation and rotation is adopted to improve the stability of pose estimation under long-range and weak-texture conditions.

\item \textbf{FW-UAV6DPose synthetic dataset:} We constructed a synthetic dataset for fixed-wing UAV 6-DoF pose estimation based on AirSim and Unreal Engine. The dataset provides accurate 6-DoF pose annotations, camera parameters, and data samples covering diverse distances, viewpoints, and pose variations, thereby supporting the training, validation, and evaluation of long-range UAV pose estimation models.
\end{itemize}

\section{Related Work}

\subsection{From Model-Based to Model-Free Pose Estimation} 
The 6-DoF pose estimation of the target has long been formulated under the assumption of instance-level model availability, where the target CAD model is known at test time for pose recovery. Representative methods such as PVNet \cite{peng2019pvnet}, HybridPose \cite{song2020hybridpose}, and GDR-Net \cite{wang2021gdr} improved instance-level pose estimation from the perspectives of keypoint voting, hybrid representations, and geometry-guided dense regression, respectively. These methods can achieve high accuracy when the target model is available. However, their strong dependence on model priors makes them difficult to apply directly to non-cooperative target scenarios where no prior information is available.

To reduce the reliance on instance-level models, category-level pose estimation has gradually attracted increasing attention. For example, NOCS established a category-shared geometric representation through a normalized object coordinate space \cite{wang2019normalized}. VI-Net \cite{lin2023vi}, RCGNet \cite{peng2022self}, and KeyPose \cite{yu2025keypose} further enhance the generalization capability of category-level pose estimation through rotation disentanglement, implicit shape modeling, and adaptive keypoint representations. Compared with instance-level methods, category-level approaches alleviate the dependence on precise CAD models to some extent. Nevertheless, they typically still rely on category priors, normalized shape spaces, or structural consistency between training and testing categories.

In recent years, model-free and few-reference 6-DoF pose estimation has further advanced the field toward generalizing to unknown objects. “Model-free” refers to the setting in which no instance-level CAD model, real-world object dimensions, or explicit 3D keypoint priors are required during inference. Gen6D \cite{liu2022gen6d} and OnePose++ \cite{he2022onepose++} demonstrated that pose recovery for novel objects can be supported by a small number of reference images or sparse-view matching. PoET \cite{jantos2023poet}, MegaPose \cite{labbe2022megapose}, FoundationPose \cite{wen2024foundationpose}, and SAM-6D \cite{lin2024sam} introduced object-centric representations, foundation-model features, rendering-guided matching, and coarse-to-fine refinement into pose estimation frameworks. Any6D \cite{lee2025any6d}, One2Any \cite{liu2025one2any}, and ZS6D \cite{ausserlechner2024zs6d} further extended pose transfer capabilities under zero-shot and few-reference settings. Overall, 6-DoF pose estimation is gradually evolving from geometric matching based on a single instance model toward object-centric representation learning and generalized inference for unknown targets.

\subsection{UAV 6-DoF Pose Estimation}

For UAV 6-DoF pose estimation, existing studies have long been dominated by the paradigm of “object detection/keypoint extraction + geometric pose solving”. Typical methods first detect the target region, then extract stable structural features such as wingtips, tail components, fuselage endpoints, anchor points, or other salient landmarks. These features are subsequently combined with camera intrinsics, prior knowledge of the target’s 3D structure, and geometric solvers such as PnP and EPnP to recover the relative pose. For example, Kim et al. adopted a combination of YOLO-based detection and PnP-based pose solving to estimate the relative pose of fixed-wing aircraft \cite{kim2021vision}. Jeong et al. enhanced keypoint localization under varying distance conditions using distance-aware keypoint heatmaps, and further recovered the 6-DoF pose of UAVs \cite{jeong2025enhanced}. DroneKey integrates temporal keypoint representations with pose-adaptive learning, thereby improving the stability of UAV keypoint detection and pose estimation in image sequences \cite{hwang2025dronekey}. In automatic landing and relative navigation tasks, pose estimation frameworks based on anchor-point detection, multi-camera collaboration, and visual–inertial fusion have also achieved promising performance \cite{tang2021learning,yu2021robust,yu2024real,tang2023n}. However, such methods generally require stably visible keypoints, accurate 2D-3D correspondences, and known target structural dimensions. Under long-range, weak-texture, and motion-blur conditions, keypoint localization errors can be further amplified by geometric pose solvers, thereby degrading the stability of pose estimation.

Beyond keypoint-based methods, some studies further exploit structural priors and geometric constraints of aircraft targets for pose recovery. For instance, geometric methods based on fuselage reference lines, wing leading-edge lines, and bilateral symmetry can estimate aircraft attitude by leveraging contour structures under weak-texture conditions \cite{teng2019pose,teng2019aircraft}. Neural-network-based feature-line extract methods combine line-feature prediction with geometric pose solving to improve robustness in complex backgrounds \cite{chen2024aircraft}. In fixed-wing relative navigation scenarios, several studies have also incorporated vision, inertial measurements, and filtering-based optimization to achieve relative state estimation under multi-source constraints \cite{izadi2015gps,ellingson2020relative,yu2021robust,yu2024real}. These methods rely on clear and stable contour structures, specific geometric assumptions, or multi-source sensor constraints. Although they are effective when sufficient conditions are available, they are difficult to satisfy the requirements of model-free, low-prior pose estimation in noncooperative long-range scenarios.

\section{Methods}

\begin{figure*}[htbp]
	\centering
	\includegraphics[width=\linewidth]{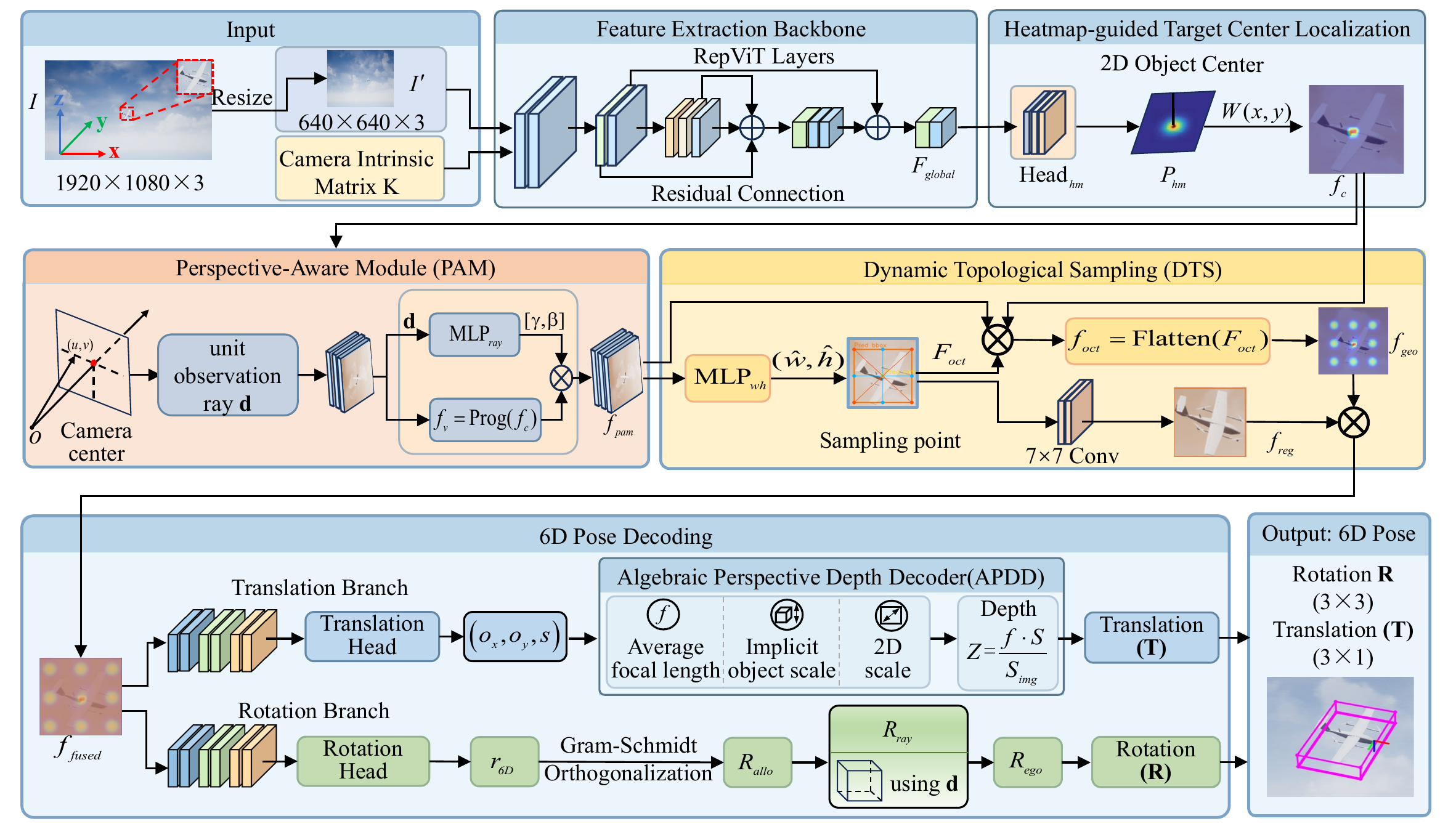}	
	\caption{Overall architecture of the proposed MF-UAVPose6D framework. It consists of heatmap-based center localization, the Perspective-Aware Module (PAM), Dynamic Topological Sampling (DTS), and a 6-DoF pose decoding module. Given a single RGB image and camera intrinsics, it performs end-to-end 6-DoF pose estimation without relying on target CAD models.}
	\label{Fig2}	
\end{figure*}

We propose MF-UAVPose6D, a model-free monocular 6-DoF pose estimation framework for long-range fixed-wing UAVs. Using only a single RGB image and camera intrinsics, the method estimates the target 6-DoF pose in an end-to-end manner without relying on a target CAD model. It achieves this through heatmap-guided target center localization, a perspective-aware module, dynamic topology sampling, and pose decoding.

\subsection{Overall Framework}

As shown in Fig. \ref{Fig2}, given an original high-resolution input image $ I \in\mathbb{R}^{H_0\times W_0\times 3}$, the proposed MF-UAVPose6D method first resizes it to a standard input resolution $I' \in \mathbb{R}^{640 \times 640 \times 3}$. A lightweight RepViT backbone is then employed to extract multi-scale visual features. Subsequently, feature maps with strides of 8, 16, and 32 are aligned to the stride-8 resolution and fused across scales, yielding a global spatial feature map:
\begin{equation}
F_{\mathrm{global}} \in \mathbb{R}^{C\times H_f\times W_f},\quad H_f=H/8,\; W_f=W/8 
\end{equation}
where $C$ denotes the number of fused feature channels, and $H,W$ are the height and width of the standard network input image $I'$.

Building upon this, the heatmap-guided target-center localization module predicts a target-center probability map $P_{\text{hm}}$ on $F_{\text{global}}$ and decodes the target center anchor $(u_c,v_c)$ together with its corresponding center feature $f_c$. PAM then incorporates the camera intrinsics $K$ to back-project the target center anchor into a unit observation ray, which is used as a geometric condition to modulate the center visual embedding, yielding the view-conditioned target representation $f_{\mathrm{pam}}$. DTS further predicts 2D bounding-box scale from $f_{\mathrm{pam}}$ and constructs an eight-point topological sampling structure around the visible center to extract structural features from edge-midpoint and corner regions on the global spatial feature map. Meanwhile, a $7\times7$ dense sampling grid is constructed on $F_{\mathrm{global}}$ according to the predicted center and 2D bounding-box scale, producing a region-level visual representation that covers the full target extent. The pose decoder takes this region-level representation as the primary appearance evidence and jointly maps the center feature $f_c$, eight-point topological features, and view-conditioned target representation $f_{\mathrm{pam}}$ into geometric context, which is injected into the translation and rotation decoding paths through branch-specific gated modulation. This produces the target-level pose representation $f_{\text{fused}}$, which integrates regional appearance, local topology, and perspective geometry constraints. Finally, the decoupled translation and rotation branches recover the 3D translation vector and rotation matrix, respectively. 

In this way, MF-UAVPose6D establishes a complete inference pipeline from target center-anchor localization to local topological modeling and geometry-consistent pose decoding, enabling monocular 6D pose estimation of long-range fixed-wing UAVs without relying on CAD model constraints.

\subsection{Heatmap-guided Target Center Localization}

To obtain a stable target center prior, we construct a lightweight heatmap prediction head on the global spatial feature map $F_{\text{global}}$. The prediction head consists of a $3\times3$ convolution, BatchNorm, ReLU, and a $1\times1$ convolution. It outputs single-channel heatmap logits, which are activated by a Sigmoid function to obtain the center probability map:

\begin{equation}
P_{\mathrm{hm}} = \mathrm{Sigmoid}\left(\mathrm{Head}_{\mathrm{hm}}(F_{\text{global}}) \right)
\end{equation}
where $P_{\mathrm{hm}}\in[0,1]^{H_f\times W_f}$ represents the probability of each feature-map location belonging to the target center, and $\mathrm{Head}_{\mathrm{hm}}(\cdot)$ denotes the heatmap prediction mapping.

To improve the stability of center decoding, the location with the maximum response is first selected as the local peak center:

\begin{equation}
(x_p, y_p) = \arg\max_{x,y} P_{\mathrm{hm}}(x, y) 
\end{equation}

A Gaussian window centered at this position is then constructed as:

\begin{equation}
G(x, y) = \exp \left( - \frac{(x - x_{\text{p}})^2 + (y - y_{\text{p}})^2}{2\sigma^2_g} \right) 
\end{equation}
where $\sigma_g$ denotes the scale of the Gaussian window and is adaptively adjusted according to the feature-map size. The target center probability map is then sharpened exponentially and multiplied element-wise with the local Gaussian window to obtain a local attention mask:

\begin{equation}
A(x, y) = \left(P_{\text{hm}}(x, y)\right)^{2.2} \cdot G(x, y) 
\end{equation}

Next, the local attention mask is normalized to obtain spatial attention weights:

\begin{equation}
W(x, y) = \frac{A(x, y)}{\sum_{x', y'} A(x', y') + \epsilon} 
\end{equation}

Based on these weights, a soft-argmax operation is employed to compute the target-center anchor: 

\begin{equation}
u = 8\sum_{x,y} W(x,y)x,\quad v = 8\sum_{x,y} W(x,y)y 
\end{equation}

Meanwhile, the network aggregates the global spatial feature map using the spatial attention weights to obtain the target-center feature: 

\begin{equation}
f_c = \sum_{x,y} W(x,y)F_{\mathrm{global}}(:,y,x) 
\end{equation}
where $f_c$ denotes the weighted visual representation of the target center region. 

This design converts the discrete heatmap peak into a continuous target center anchor estimate. While preserving the global detection capability, it improves the sub-pixel stability of small-target center localization.

\subsection{Perspective-Aware Module}

In monocular images, the same object located at different image positions corresponds to different observation rays, leading to perspective variations induced by changes in the viewing angle. If the network relies solely on the visual response of the target center region, it may struggle to distinguish perspective changes caused by spatial displacement from the object's true 3D rotation \cite{mousavian20173d,liu2018intriguing}. Therefore, we design a perspective-aware module, which constructs an observation-direction prior using the camera intrinsics and the target-center location, and injects it into the target-center feature as a geometric condition to enhance the view-aware representation of the target. 

Specifically, given the target center anchor $(u,v)$ predicted by the heatmap-guided target center localization module and the camera intrinsic matrix $K$, the anchor is back-projected into the normalized camera coordinate system to obtain a unit observation ray:

\begin{equation}
\mathbf{d} = \frac{K^{-1}[u, v, 1]^T}{\|K^{-1}[u, v, 1]^T\|_2} 
\end{equation}
where the unit observation ray $\mathbf{d}$ denotes the observation direction of the target center with respect to the camera optical center, encoding the viewing-angle information of the target in the current field of view.

The $\mathbf{d}$ is then fed into a ray encoder, $\mathrm{MLP}_{\mathrm{ray}}$, to generate channel-wise modulation parameters: 

\begin{equation}
\left[\gamma,\beta\right] = \mathrm{MLP}_{\mathrm{ray}}( \mathbf{d}) 
\end{equation}
where $\gamma$ and $\beta$ denote the channel-wise scaling and shifting parameters derived from the $\mathbf{d}$, respectively. Meanwhile, the target center feature $f_c$ is linearly projected to obtain a visual embedding:

\begin{equation}
f_v = \mathrm{Proj}(f_c) 
\end{equation}
where $\mathrm{Proj}(\cdot)$ denotes the projection function that maps the target center feature into the hidden feature space. Finally, PAM injects the observation-direction prior information into the visual embedding through channel-wise conditional modulation:

\begin{equation}
f_{\mathrm{pam}} = f_v \odot \mathrm{Sigmoid}(\gamma) + \beta 
\end{equation}
where $\odot$ denotes channel-wise multiplication, and $f_{\mathrm{pam}}$ is the view-conditioned target representation after perspective-aware modulation.

Compared with the original center feature $f_c$, $f_{\mathrm{pam}}$ jointly encodes the appearance response of the target-center region and the observation direction prior. It therefore provides viewpoint-aware geometric constraints for subsequent 2D scale prediction, dynamic topological sampling, and pose decoding.

\subsection{Dynamic Topological Sampling}

Fixed-wing UAVs exhibit a pronounced elongated structure, and their pose estimation heavily relies on geometric cues from the wings, tail, fuselage edges, and other structural components. Using only local features around the target center may fail to capture these critical structures, especially when the target is small or undergoes large viewpoint variations. To address this issue, we design a DTS module, which constructs an adaptive eight-point sampling structure according to the predicted target scale and extracts structural context from the target edge and corner regions. 

First, DTS takes the perspective-aware feature $f_{\mathrm{pam}}$ as input and estimates the 2D bounding-box scale of the target in the image plane through a width-height prediction branch:
\begin{equation}
(\hat{w},\hat{h}) = \mathbf{a} \cdot \exp\left(\mathrm{MLP}_{\mathrm{wh}}(f_{\mathrm{pam}})\right) 
\end{equation}
where $\mathrm{MLP}_{\mathrm{wh}}(\cdot)$ denotes the width-height prediction branch, and $\mathbf{a}=(0.1W,0.1H)$ is the base scale anchor. The network predicts logarithmic scale offsets, which are mapped by $\exp(\cdot)$ to obtain positive 2D bounding-box dimensions.

Then, taking the target center anchor $(u,v)$ as the origin, an eight-point topological sampling structure is constructed based on the predicted scale $(\hat{w},\hat{h})$. This structure consists of four edge midpoints and four corner points of the target bounding box. The relative offset matrix is defined as:
\begin{equation}
\Delta P = \frac{1}{2}
\begin{bmatrix}
\hat{w} & 0 \\
-\hat{w} & 0 \\
0 & \hat{h} \\
0 & -\hat{h} \\
\hat{w} & \hat{h} \\
-\hat{w} & \hat{h} \\
\hat{w} & -\hat{h} \\
-\hat{w} & -\hat{h}
\end{bmatrix} 
\end{equation}

The dynamic sampling points in the image coordinate system are therefore obtained as:
\begin{equation}
P_{\text{oct}} = (u_c, v_c) + \Delta P 
\end{equation}
where $P_{\mathrm{oct}}\in\mathbb{R}^{8\times2}$ denotes the eight dynamic sampling points. 

These points are further normalized to the $[-1,1]$ sampling coordinate system, and bilinear sampling is performed on $F_{\text{global}}$ to extract the eight-point topological features:
\begin{equation}
F_{\mathrm{oct}} = \mathrm{GridSample}(F_{\mathrm{global}}, P_{\mathrm{oct}}) 
\end{equation}

The sampled features are then flattened into a structural feature vector:
\begin{equation}
f_{\mathrm{oct}} = \mathrm{Flatten}(F_{\text{oct}}) 
\end{equation}

Finally, the network concatenates $f_c$, $f_{\text{oct}}$, and $f_{\text{pam}}$ along the channel dimension to obtain a unified target-level pose representation:
\begin{equation}
f_{\text{fused}} = \mathrm{Concat}( f_{\text{c}}, \; f_{\text{oct}}, \; f_{\text{pam}}) 
\end{equation}

Through this design, DTS extends a single center-point representation into a structured representation covering the target center, edges, and corners. Without introducing an explicit CAD model, it supplements the edge and contour cues of fixed-wing UAVs, thereby providing more stable local geometric constraints for subsequent translation and rotation decoding.

The eight-point sampling features are then flattened into a structural feature vector to explicitly represent the sparse topology around the target boundary:
\begin{equation}
f_{\mathrm{oct}} = \mathrm{Flatten}(F_{\mathrm{oct}})
\end{equation}

Meanwhile, using the predicted center $(u_c,v_c)$ and scale $(\hat{w},\hat{h})$ as geometric anchors, the model constructs a $7\times7$ dense sampling grid on the global spatial feature map to extract a region-level visual representation $f_{\mathrm{reg}}$ that covers the predicted target area. The center feature, eight-point topological feature, and PAM feature are further concatenated to form the geometric context feature $f_{\mathrm{geo}}$:
\begin{equation}
f_{\mathrm{geo}}=\mathrm{Concat}\left(f_c,f_{\mathrm{oct}},f_{\mathrm{pam}}\right)
\end{equation}

Then, $f_{\mathrm{reg}}$ and $f_{\mathrm{geo}}$ are adaptively fused to obtain the target-level pose representation $f_{\mathrm{fused}}$:
\begin{equation}
f_{\mathrm{fused}}=\Gamma\left(f_{\mathrm{reg}},\Phi^g(f_{\mathrm{geo}})\right)
\end{equation}
where $\Phi^g(\cdot)$ denotes the geometric context projection, and $\Gamma(\cdot)$ represents the gated fusion function. Thus, the model obtains a stable pose representation that integrates holistic target features with geometric constraints.

\subsection{Pose Decoding}

After obtaining the pose representation $f_{\text{fused}}$, directly regressing rotation and translation with shared fully connected layers may introduce multi-task feature interference. This is because translation estimation is highly sensitive to target scale, center offset, and depth variation, whereas rotation estimation relies more on target shape structure and viewpoint changes. Therefore, we adopt a decoupled branch design, where $f_{\text{fused}}$ is fed into the translation and rotation branches separately: 
\begin{equation}
f_{\mathrm{trans}} = \mathrm{MLP}_t(f_{\mathrm{fused}}),\quad f_{\mathrm{rot}} = \mathrm{MLP}_r(f_{\mathrm{fused}}) 
\end{equation}
where $\mathrm{MLP}_{t}(\cdot)$ and $\mathrm{MLP}_{r}(\cdot)$ denote the translation and rotation decoding branches, respectively. The two branches do not share parameters and learn task-specific feature mappings for translation and rotation estimation.

\subsubsection{\textbf{Translation Branch and Algebraic Perspective Depth Decoder}}

Translation estimation requires recovering both the in-plane center offset and the depth along the optical axis. For long-range fixed-wing UAVs, the scale variation of the target in the image is often subtle. If the absolute depth $Z$ is directly regressed, the network has to learn a large-range continuous distance mapping from limited appearance variations, which can easily lead to scale drift. To address this issue, we introduce an algebraic perspective depth decoder (APDD) in the translation branch, which reformulates depth estimation from direct numerical regression into a geometry-constrained decoding process based on the pinhole imaging scale relationship. 

Specifically, the translation branch takes $f_{\mathrm{trans}}$ as input and predicts three intermediate variables through a translation prediction head $\mathrm{Head}_{\mathrm{trans}}(\cdot)$: 

\begin{equation}
\left(o_x,o_y,s\right) = \mathrm{Head}_{\mathrm{trans}}(f_{\mathrm{trans}}) 
\end{equation}
here, $o_x$ and $o_y$ denote the normalized 2D origin-offset predictions, while $s$ represents the logarithmic prediction of the implicit scale. The model then converts $(o_x,o_y)$ into local origin-projection refinements in pixel coordinates:
\begin{equation}
\Delta u = \lambda_o\hat{w}\tanh(o_x),\quad \Delta v = \lambda_o\hat{h}\tanh(o_y)
\end{equation}
where $\Delta u$ and $\Delta v$ denote the horizontal and vertical pixel offsets relative to the target center anchor $(u_c,v_c)$, respectively. $\lambda_o$ is the origin-offset range coefficient, which is set to $\lambda_o=1.25$ in the current implementation, enabling the translation branch to perform stable geometric refinement within the local target region.

For the depth estimation, APDD does not directly output the absolute depth. Instead, it maps $s$ to a positive implicit physical scale through an exponential function: 
\begin{equation}
S=\exp(s) 
\end{equation}
it should be emphasized that (S) is neither a predefined scale nor a ground-truth dimension derived from the CAD model; rather, it is an implicit scale learned through supervised training and adaptively predicted by the network for each input.

Given the 2D bounding-box scale $(\hat{w},\hat{h})$ predicted by the DTS module, the apparent image scale of the target is defined as follows:
\begin{equation}
S_{\mathrm{img}} = \sqrt{\hat{w}\hat{h}+\epsilon} 
\end{equation}

The average focal length is computed as:
\begin{equation}
f = \frac{f_x + f_y}{2} 
\end{equation}

According to the inverse relationship among physical scale, image scale, and depth in the pinhole camera model, APDD decodes the depth as
\begin{equation}
Z = \frac{f \cdot S}{S_{\mathrm{img}}} 
\end{equation}

In practice, $Z$ is constrained within the valid depth range of the dataset, $[Z_{\min},Z_{\max}]$, to avoid invalid depth values caused by abnormal scale predictions. With this design, the network only needs to learn an implicit scale variable related to the target instance, pose, and visible size, while depth recovery is performed through explicit perspective geometry. This reduces the ill-posedness of long-range monocular depth regression. 

After obtaining the depth $Z$, the 3D translation is recovered through perspective back-projection: 

\begin{equation}
t_x = \frac{(u+\Delta u-c_x)Z}{f_x},\quad t_y = \frac{(v+\Delta v-c_y)Z}{f_y} 
\end{equation}

The final 3D translation vector in the camera coordinate system is given by
\begin{equation}
T = \left[t_x,t_y,Z\right]^\mathrm{T} 
\end{equation}

This decoding strategy transforms the challenging direct regression of absolute depth into an algebraic inference process jointly constrained by the camera focal length, the 2D apparent scale, and the implicit physical scale. Compared with directly regressing $Z$, APDD explicitly exploits the scale–depth relationship inherent in monocular imaging, thereby maintaining geometric consistency between depth estimation and 3D back-projection. It is worth noting that the implicit scale $S$ is learned by the network from supervised training data, rather than being provided by any explicit 3D model, real-world object dimensions, or 3D keypoint prior.

\subsubsection{\textbf{Rotation Branch and Ray-Aligned Decoding}}

The rotation branch takes $f_{\mathrm{rot}}$ as input and outputs a 6D continuous rotation representation through a rotation prediction head:

\begin{equation}
\mathbf{r}_{\mathrm{6D}}=\mathrm{Head}_{\mathrm{rot}}(f_{\mathrm{rot}}) 
\end{equation}
where $\mathrm{Head}_{\mathrm{rot}}(\cdot)$ denotes the rotation prediction head, and $\mathbf{r}_{\mathrm{6D}}\in\mathbb{R}^{6}$ represents a continuous rotation representation composed of two 3D vectors. Compared with Euler angles and quaternions, the 6D rotation representation avoids angular singularities and sign discontinuities, making it more suitable for neural network regression.

To obtain a valid rotation matrix, the network first applies Gram-Schmidt orthogonalization to $\mathbf{r}_{\mathrm{6D}}$, converting it into an allocentric rotation matrix:
\begin{equation}
R_{\mathrm{allo}} = \mathrm{GS}\left(\mathbf{r}_{\mathrm{6D}}\right) 
\end{equation}
where $\mathrm{GS}(\cdot)$ denotes the Gram-Schmidt orthogonalization operation. $R_{\mathrm{allo}}$ represents the rotation of the target with respect to the current observation direction, focusing on the relative pose under the viewing ray rather than directly regressing the final rotation in the camera coordinate system. 

To transform this relative pose into the camera coordinate system, we use the unit observation ray $\mathbf{d}$ obtained in PAM to construct a ray rotation matrix $R_{\mathrm{ray}}$. This matrix represents the rotation from the camera principal-axis direction to the target observation-ray direction. The final egocentric rotation matrix in the camera coordinate system is obtained by
\begin{equation}
R_{\text{ego}} = R_{\text{ray}} \cdot R_{\text{allo}} 
\end{equation}
where $R_{\mathrm{ego}}$ is the final predicted rotation matrix of the target.

This rotation decoding strategy explicitly models the viewing-angle variation induced by the target image position through a geometric transformation. As a result, the rotation branch mainly learns the mapping between the target appearance structure and its relative pose, which reduces the difficulty of directly regressing camera-coordinate rotation and alleviates the perspective coupling caused by image-position changes.

\subsection{Joint Loss Function}

MF-UAVPose6D is trained with end-to-end joint supervision. The overall loss consists of the heatmap detection loss $\mathcal{L}_{\text{hm}}$, 2D bounding-box scale loss $\mathcal{L}_{\text{wh}}$, 3D translation loss $\mathcal{L}_{\text{trans}}$, and 3D rotation loss $\mathcal{L}_{\text{rot}}$:

\begin{equation}
\mathcal{L}_{\text{total}} = \lambda_1 \mathcal{L}_{\text{hm}} + \lambda_2 \mathcal{L}_{\text{wh}} + \lambda_3 \mathcal{L}_{\text{trans}} + \lambda_4\mathcal{L}_{\text{rot}} 
\end{equation}
where $\lambda_1,\lambda_2,\lambda_3,\lambda_4$ are balancing weights for the corresponding loss terms, and are set to 50, 4, 4, and 25, respectively.

For the heatmap branch, we adopt the penalty-reduced focal loss to mitigate the imbalance between sparse target-center regions and dominant background regions:

\begin{equation}
\mathcal{L}_{\mathrm{hm}} = -\frac{1}{N}\sum_{x,y}
\begin{cases}
(1-\hat{P}_{xy})^2\log(\hat{P}_{xy}), & P_{xy}=1,\\
(1-P_{xy})^4\hat{P}_{xy}^2\log(1-\hat{P}_{xy}), & P_{xy}<1 
\end{cases}
\end{equation}
where $\hat{P}_{xy}$ and $P_{xy}$ denote the predicted and ground-truth heatmap responses at location $(x,y)$, respectively, and $N$ is the number of positive samples.

For the 2D bounding-box branch, a Smooth L1 loss is adopted:

\begin{equation}
\mathcal{L}_{\mathrm{wh}} = \mathrm{SmoothL1}\big((\hat{w},\hat{h}),(w,h)\big) 
\end{equation}
where $(\hat{w},\hat{h})$ and $(w,h)$ denote the predicted and ground-truth bounding-box sizes, respectively.

For the 3D translation branch, considering the large absolute depth range in long-distance scenarios, the translation loss is decomposed into a logarithmic depth loss and a depth-normalized planar translation loss:

\begin{equation}
\mathcal{L}_{\mathrm{trans}} = \mu_1\,\mathrm{SmoothL1}(\log \hat{Z}, \log Z) + \mu_2\,\frac{\mathrm{SmoothL1}(\hat{\mathbf{T}}_{xy},\mathbf{T}_{xy})}{|Z|+\epsilon} 
\end{equation}
where $\hat{Z}$ and $Z$ denote the predicted and ground-truth depths, respectively, and $\hat{\mathbf{T}}_{xy}$ and $\mathbf{T}_{xy}$ denote the predicted and ground-truth lateral translation components. $\mu_1$ and $\mu_2$ are balancing coefficients within the translation loss, and are set to 20 and 50, respectively.

For the 3D rotation branch, we use the SO(3) geodesic loss $\mathcal{L}_{\mathrm{geo}}$ as the primary supervision and introduce a Smooth L1 regularization term on the 6D representation:

\begin{equation}
\mathcal{L}_{\mathrm{rot}} = \mathcal{L}_{\mathrm{geo}} + \eta \mathcal{L}_{\mathrm{6D}} 
\end{equation}
where

\begin{equation}
\mathcal{L}_{\mathrm{geo}} = \frac{1}{\pi} \arccos\left( \frac{\mathrm{Tr}\left( \hat{R} R^T \right) - 1}{2} \right) 
\end{equation}

\begin{equation}
\mathcal{L}_{\mathrm{6D}} = \mathrm{SmoothL1}\left( \hat{\mathbf{r}}_{\mathrm{6D}}, \mathbf{r}_{\mathrm{6D}} \right) 
\end{equation}
here, $\hat{R}$ and $R$ denote the predicted and ground-truth rotation matrices, respectively, while $\hat{\mathbf{r}}_{\text{6D}}$ and $\mathbf{r}_{\text{6D}}$ denote the predicted and ground-truth 6D continuous rotation representations. The weight of the 6D regularization term is set to $\eta=5$.

Through the above joint loss design, the network simultaneously optimizes 2D center localization, target apparent scale estimation, 3D translation, and manifold-aware rotation, thereby establishing a consistent 2D-3D pose learning constraint.

\section{Experiment}
\subsection{Dataset}

This paper constructed the FW-UAV6DPose synthetic dataset based on AirSim and Unreal Engine for long-range fixed-wing UAV 6-DoF pose estimation. Unlike conventional 6-DoF pose datasets that predominantly focus on close-range, texture-rich objects, FW-UAV6DPose is designed for hundred-meter-scale aerial observation scenarios. It emphasizes the distinctive characteristics of fixed-wing UAVs under long-range imaging conditions, including small size, slender structure, weak texture, pronounced scale variation, and strong background interference, thereby better aligning with practical requirements in aerial surveillance, interception early warning, and relative navigation. The data acquisition process is conducted in a simulated environment, where the UAV's autonomous flight is modeled using its inherent flight dynamics to generate corresponding body attitudes. Meanwhile, the camera pose and viewing angle are adaptively adjusted to ensure that the target remains within the effective field of view, thereby capturing UAV pose observations from the camera perspective. This process yields a dataset spanning multiple observation distances, viewpoints, and attitudes, providing reliable data to support UAV 6-DoF pose estimation. The dataset is available online\footnote{The dataset is available online: \url{https://1drv.ms/f/c/29da00c199018e43/IgDeb1nx1EPVTpUCrTQETU04AdkbyH2DbYoCyVI7EHKDHEw?e=cZjeZT}}.

\begin{figure}[!t]
	\centering
	\includegraphics[width=\linewidth]{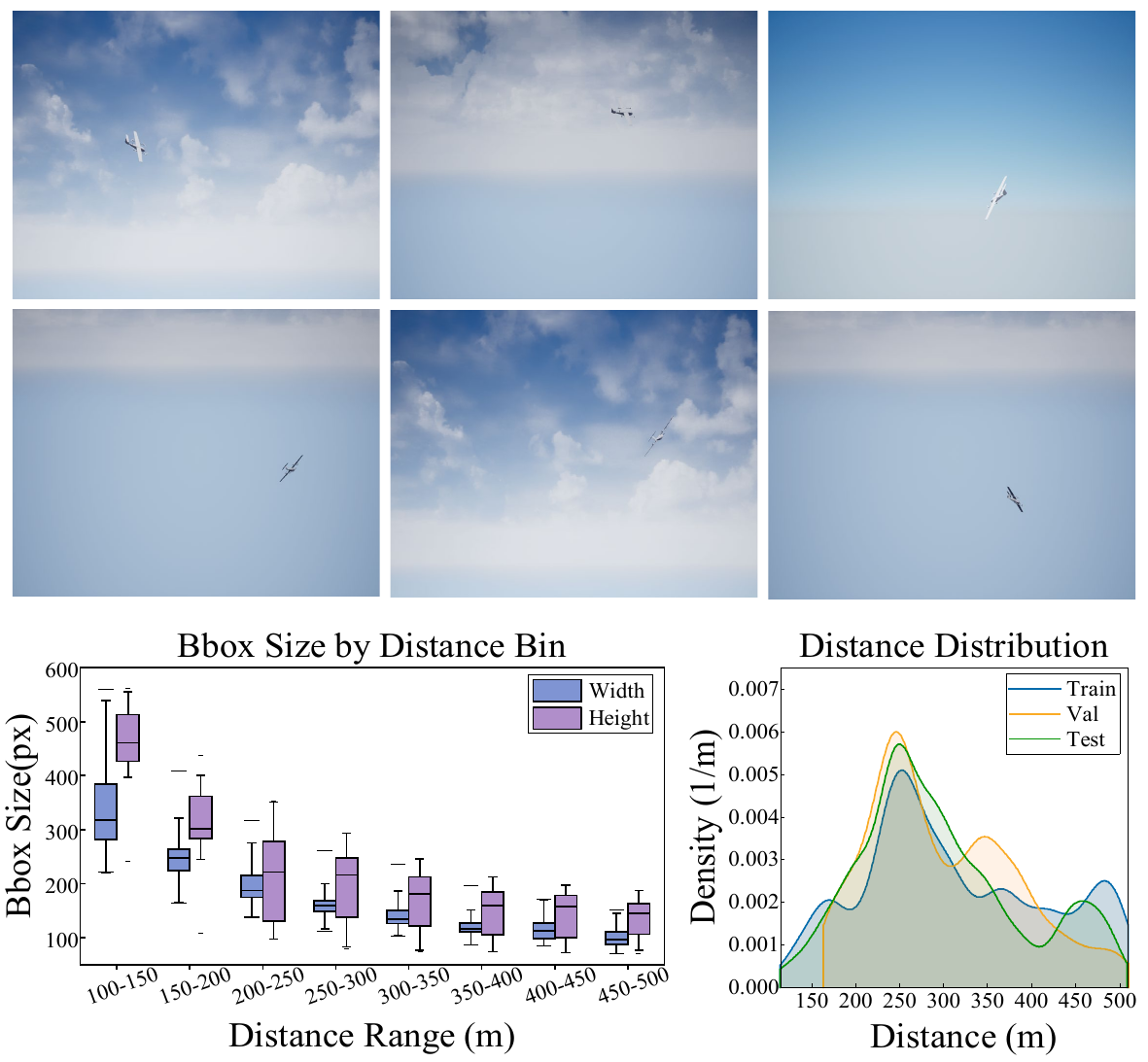}	
	\caption{Examples and statistical distributions of the FW-UAV6DPose dataset. The upper panels show fixed-wing UAV samples under different scenarios, while the lower panels present the statistical distribution of the dataset.
}
	\label{dataset}	
\end{figure}

The dataset is organized in the BOP format and includes RGB images, visible-region masks, camera intrinsics, 2D visible bounding boxes, and ground-truth 6-DoF pose of the target with respect to the camera coordinate system. The image resolution is $1920\times1080$. FW-UAV6DPose contains 161 sequences and 15,325 annotated images in total, with 7725, 2400, and 4500 images assigned to the training, validation, and test sets, respectively. The target distance ranges from approximately 100 to 500m, and the average visible bounding-box area accounts for only about 1.20\% of the full image. Fig. \ref{dataset} presents representative samples from the FW-UAV6DPose dataset and its statistical distributions. The dataset is available online.


\subsection{Evaluation Metrics}

This study evaluates the proposed method from three complementary perspectives: rotational accuracy, translational accuracy, and overall 6-DoF pose consistency. Let the predicted pose be denoted as $(\hat{R}, \hat{T})$ and the ground-truth pose as $(R, T)$, where $R, \hat{R} \in SO(3)$ represent rotation matrices, and $T, \hat{T} \in \mathbb{R}^{3}$ denote translation vectors expressed in the camera coordinate system.

The rotation error measures the angular deviation between the predicted rotation and the ground-truth rotation. In this work, the rotation error is defined using the geodesic distance on $SO(3)$:
\begin{equation}
E_r = \arccos\left(\frac{\operatorname{Tr}(\hat{R}R^{T}) - 1}{2}\right)\cdot\frac{180}{\pi}
\end{equation}
where $\operatorname{Tr}(\cdot)$ denotes the matrix trace operator. This metric represents the minimum angular discrepancy between two three-dimensional rotations, measured in degrees. A smaller value indicates more accurate attitude estimation.

The translation error quantifies the three-dimensional distance deviation between the predicted object position and the ground-truth object position. In this work, it is defined as the Euclidean distance:
\begin{equation}
E_t = \left\|\hat{T} - T\right\|_2
\end{equation}
This metric jointly reflects both lateral localization error and depth estimation error. Since the dataset considered in this study targets long-range UAV observation scenarios, the translation error is reported in meters. A smaller value indicates more accurate estimation of the target’s relative position.

In addition, we adopt Pose@10$^\circ$/5\% to evaluate the joint pose estimation capability when both rotation and translation satisfy the required accuracy thresholds. It denotes the proportion of samples for which the predicted pose has a rotation error below 10$^\circ$ and a relative translation error below 5\%.

To further evaluate the overall 6-DoF pose accuracy under the combined effects of rotation and translation, this study adopts the Average Distance of Model Points (ADD) metric. Given a set of target model points $\mathcal{M}$, ADD is defined as the mean Euclidean distance between corresponding 3D points transformed by the predicted pose and the ground-truth pose:
\begin{equation}
ADD =\frac{1}{|\mathcal{M}|}\sum_{\mathbf{x}\in\mathcal{M}}\|(\hat{R}\mathbf{x}+\hat{T})-(R\mathbf{x}+T)\|_2
\end{equation}
Compared with rotation error and translation error considered independently, ADD provides a more direct measure of the integrated impact of the complete 6-DoF pose on the spatial alignment of the target’s three-dimensional structure.

\subsection{Comparison with Prior Works}

\begin{table*}[!t]
\centering
\caption{Comparison of 6-DoF pose Estimation Methods on the FW-UAV6DPose Dataset}
\label{taball}
\small
\setlength{\tabcolsep}{3pt}
\resizebox{\linewidth}{!}{
\begin{tabular}{l|c|c|c|c|c|c|c}
\hline
Method & CAD Required & Inputs Required for Inference & $E_r$ ($\downarrow$) & $E_t$ ($\downarrow$) & Pose@10°/5\% ($\uparrow$) & ADD-0.5d ($\uparrow$) & Runtime ($\downarrow$) \\
\hline
DronePose \cite{albanis2020dronepose}      & Yes & RGB, camera intrinsics, CAD & 29.48$^\circ$ & 28.89m & 43.1\% & 44.6\% & 2.3 ms \\
\hline
GDR-Net \cite{wang2021gdr} & Yes & \makecell[c]{RGB, camera intrinsics, \\CAD model, 2D bounding box} & 9.15° & 7.9m & 74.62\% & 83.5\% & 6.8 ms \\
\hline
PoET \cite{jantos2023poet}          & No  & RGB, 2D bounding box & 45.24$^\circ$ & 25.29m  & 24.3\% & 54.2\% & 106.2ms \\
\hline
SC6D-Pose \cite{cai2022sc6d}     & No  & \makecell[c]{RGB, camera intrinsics, \\2D bounding box} & 9.36$^\circ$ & 3.26m  & 73.7\% & 99.7\% & 13.2ms \\
\hline
MF-UAVPose6D (ours)           & No  & RGB, camera intrinsics & 4.96$^\circ$ & 7.28m  & 82.8\% & 88.5\% & 4.6ms \\
\hline
\end{tabular}
}
\end{table*}

\begin{figure*}[!t]
\centering
\begin{subfigure}[t]{0.24\textwidth}
    \centering
    \includegraphics[width=\linewidth]{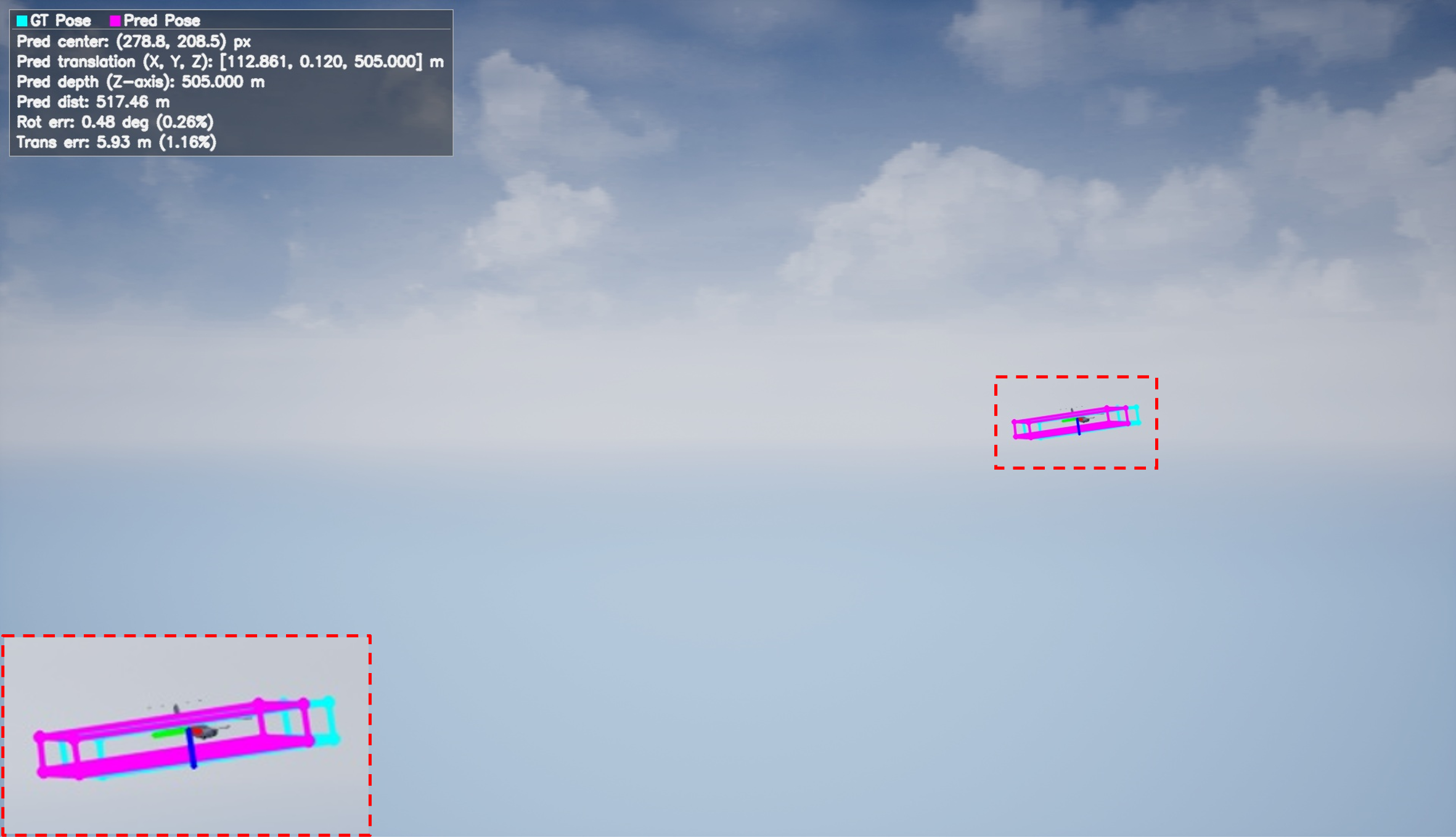}
\end{subfigure}
\hfill
\begin{subfigure}[t] {0.24\textwidth}
    \centering
    \includegraphics[width=\linewidth]{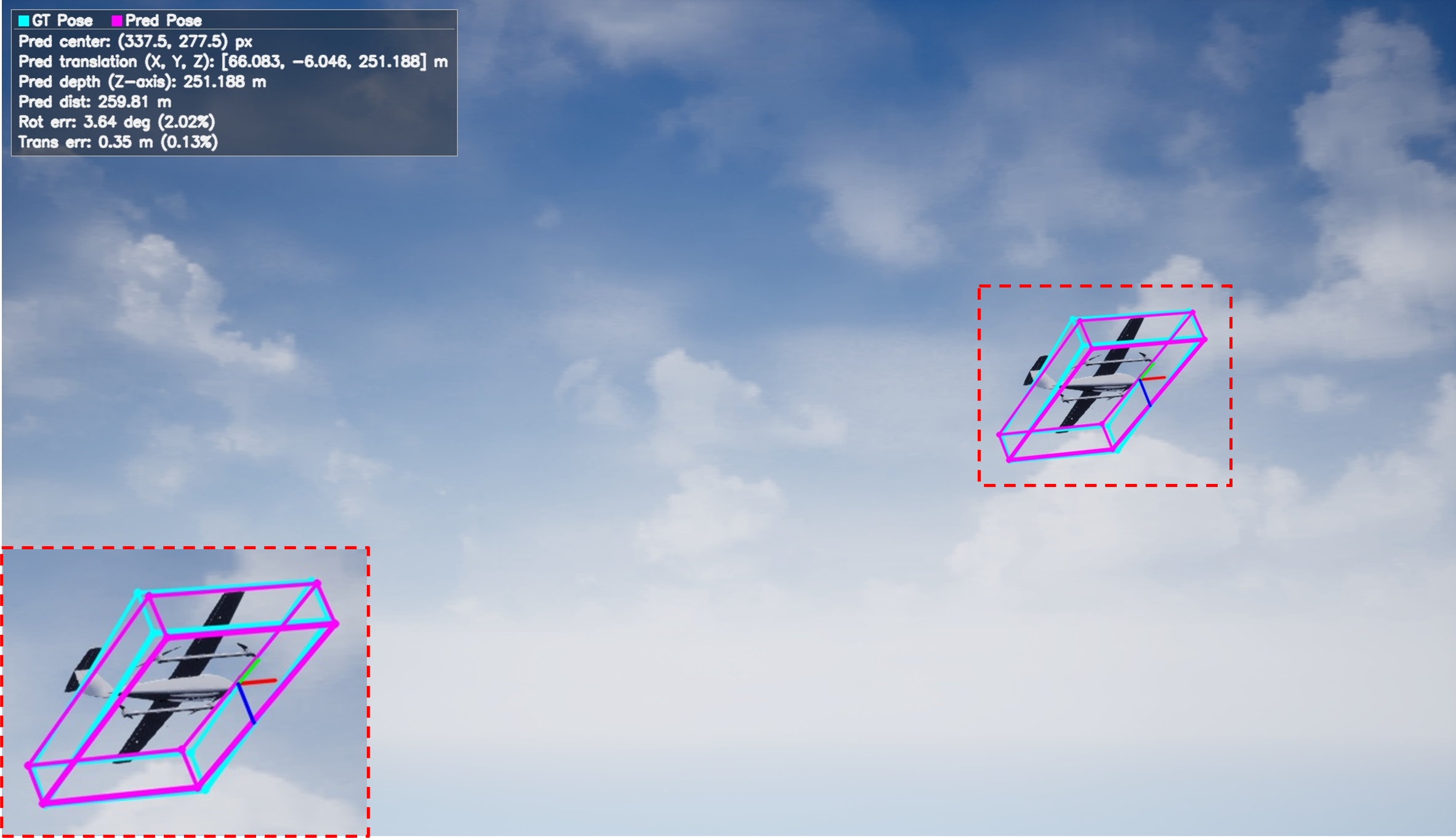}
    
\end{subfigure}
\hfill
\begin{subfigure}[t]{0.24\textwidth}
    \centering
    \includegraphics[width=\linewidth]{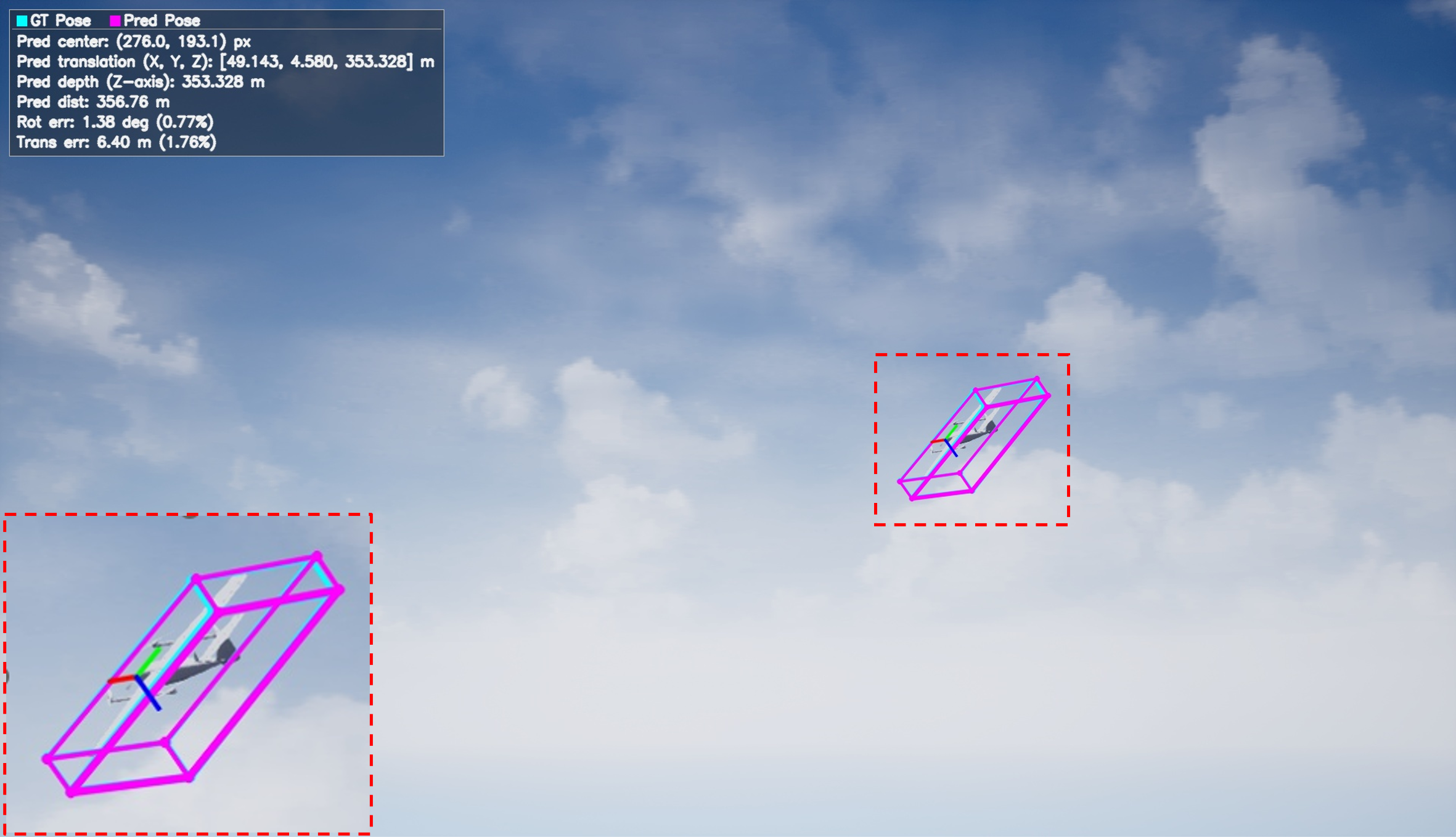}
    
\end{subfigure}
\hfill
\begin{subfigure}[t]{0.24\textwidth}
    \centering
    \includegraphics[width=\linewidth]{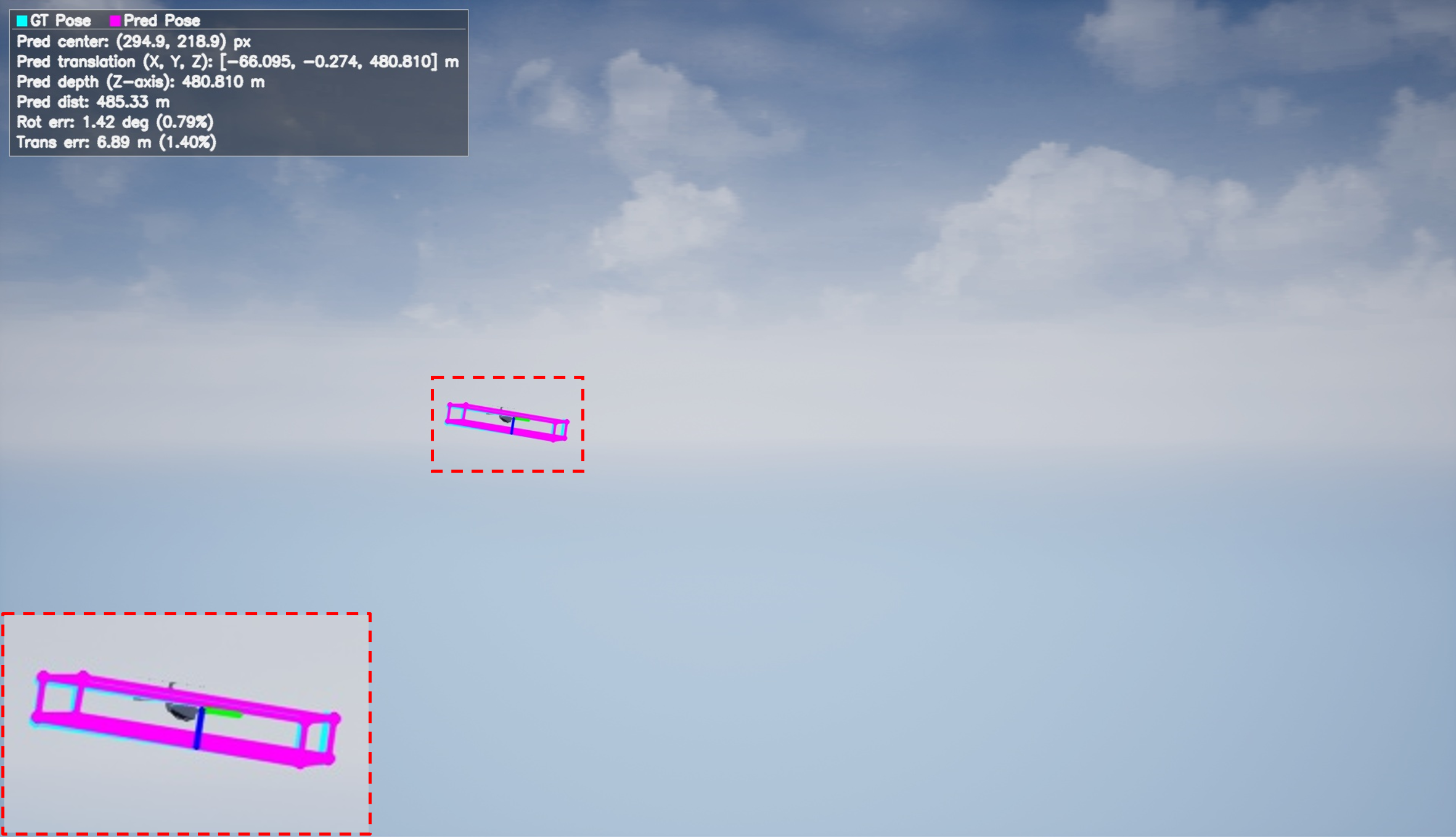}
\end{subfigure}

\vspace{0.5em}

\begin{subfigure}[t]{0.24\textwidth}
    \centering
    \includegraphics[width=\linewidth]{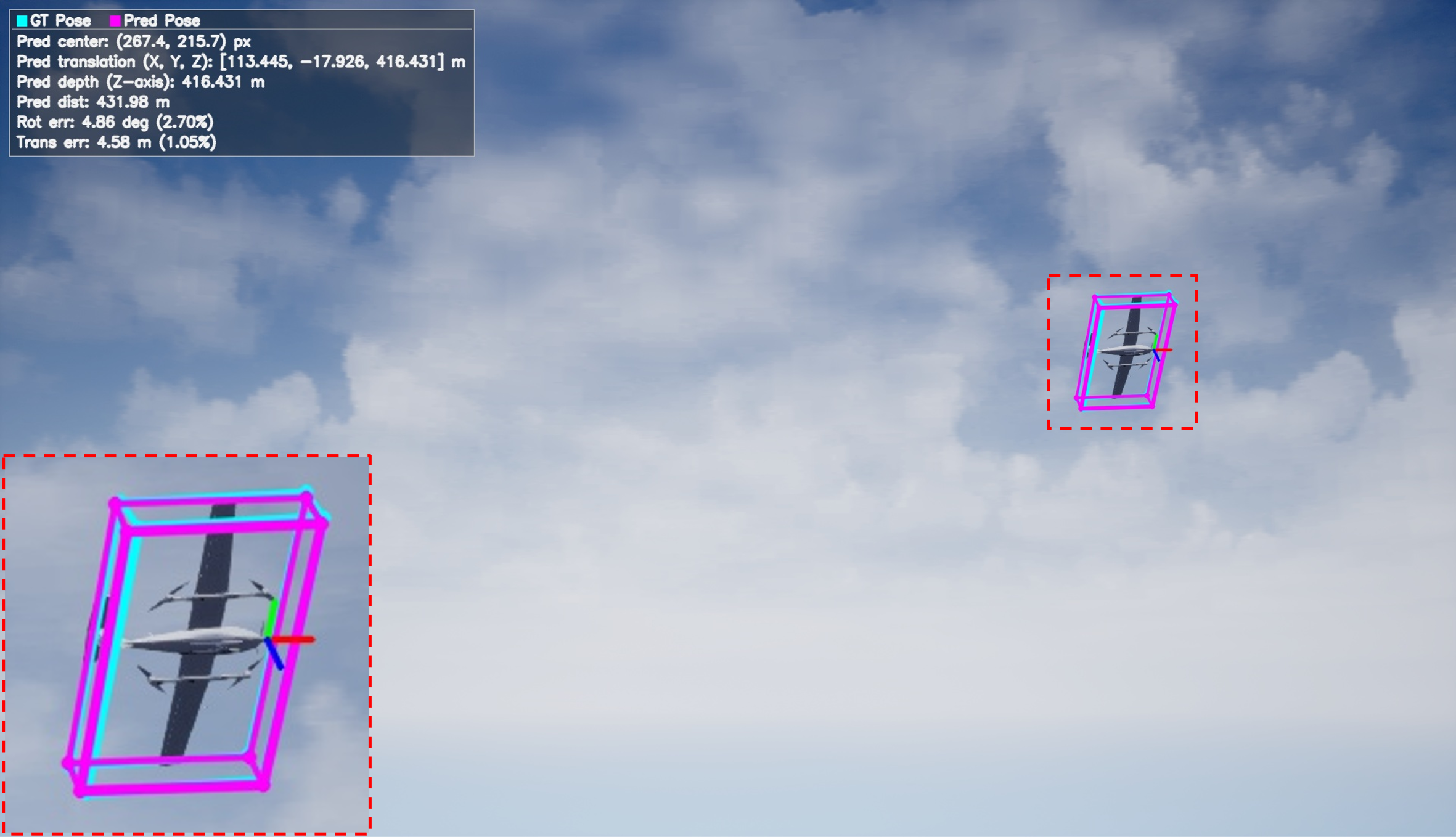}
\end{subfigure}
\hfill
\begin{subfigure}[t]{0.24\textwidth}
    \centering
    \includegraphics[width=\linewidth]{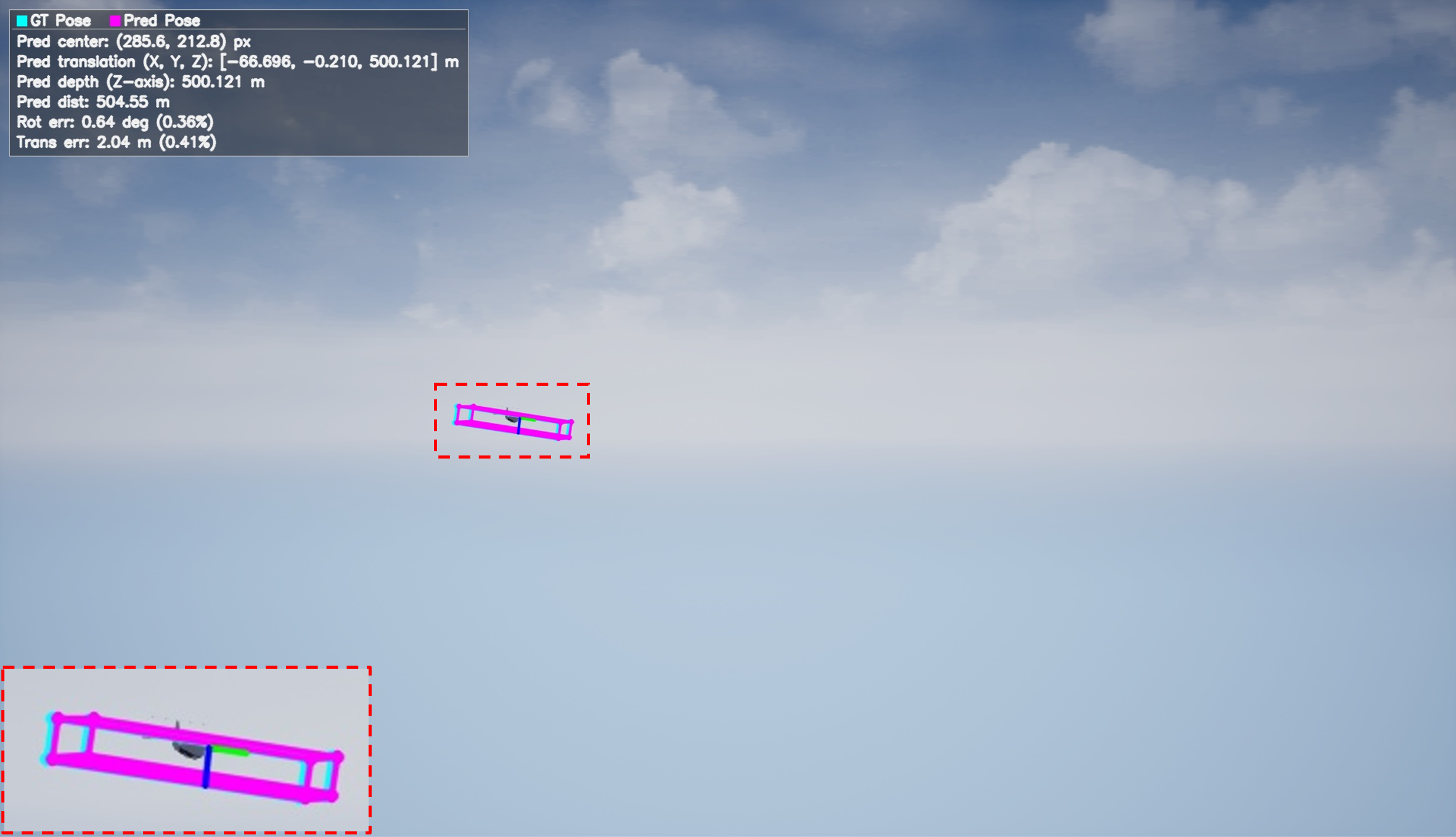}
\end{subfigure}
\hfill
\begin{subfigure}[t]{0.24\textwidth}
    \centering
    \includegraphics[width=\linewidth]{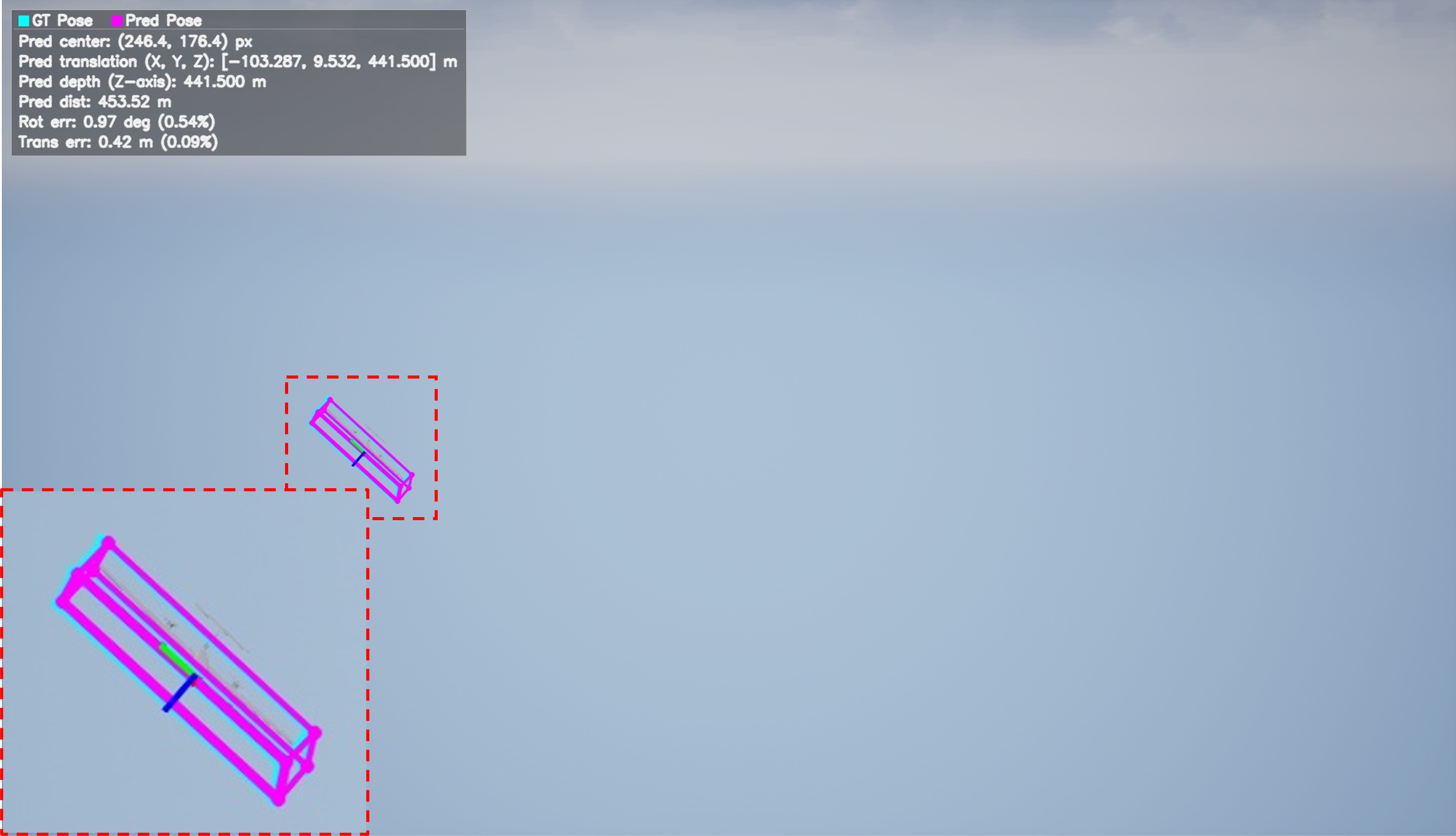}
\end{subfigure}
\hfill
\begin{subfigure}[t]{0.24\textwidth}
    \centering
    \includegraphics[width=\linewidth]{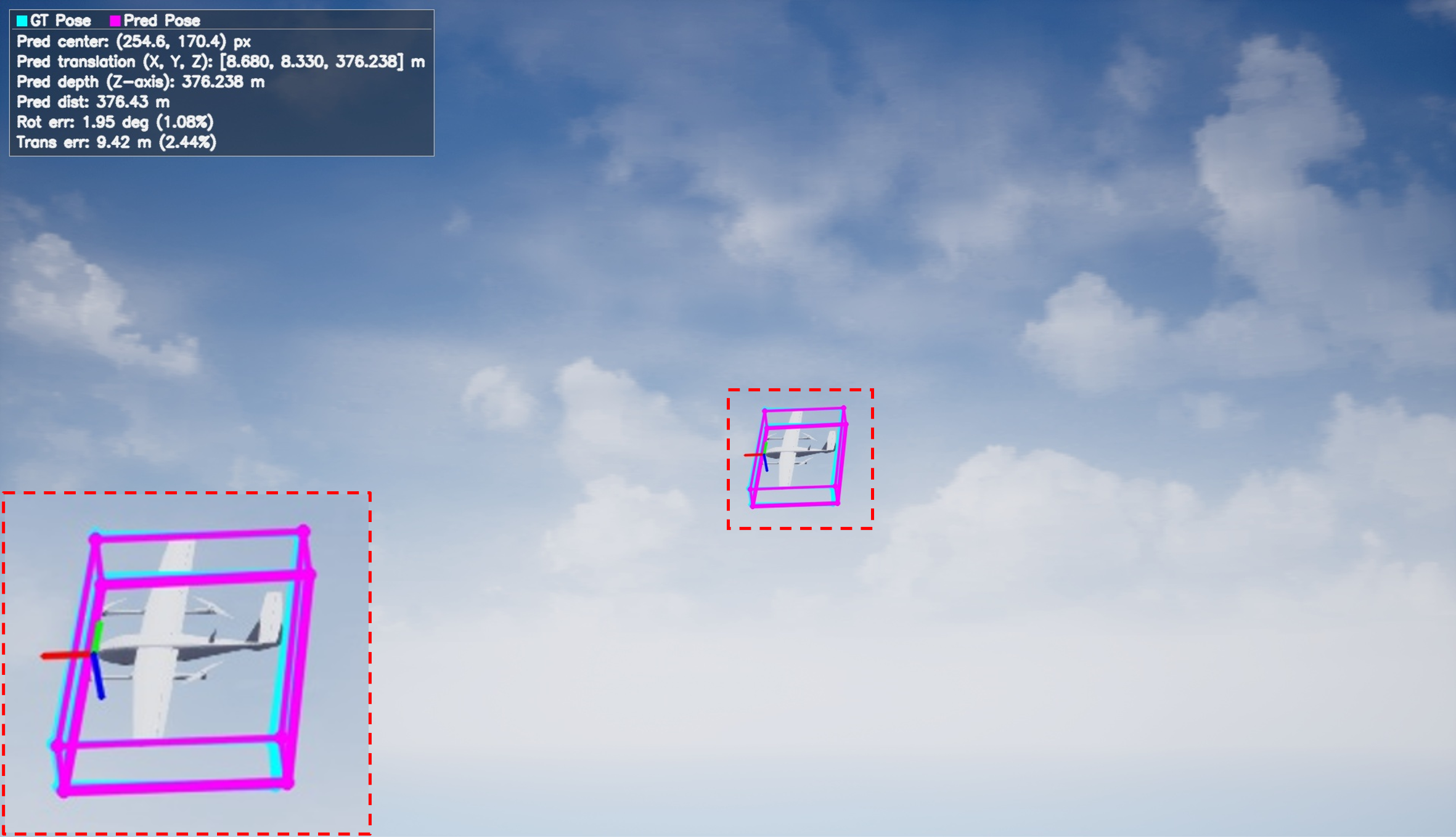}
\end{subfigure}

\vspace{0.5em}

\begin{subfigure}[t]{0.24\textwidth}
    \centering
    \includegraphics[width=\linewidth]{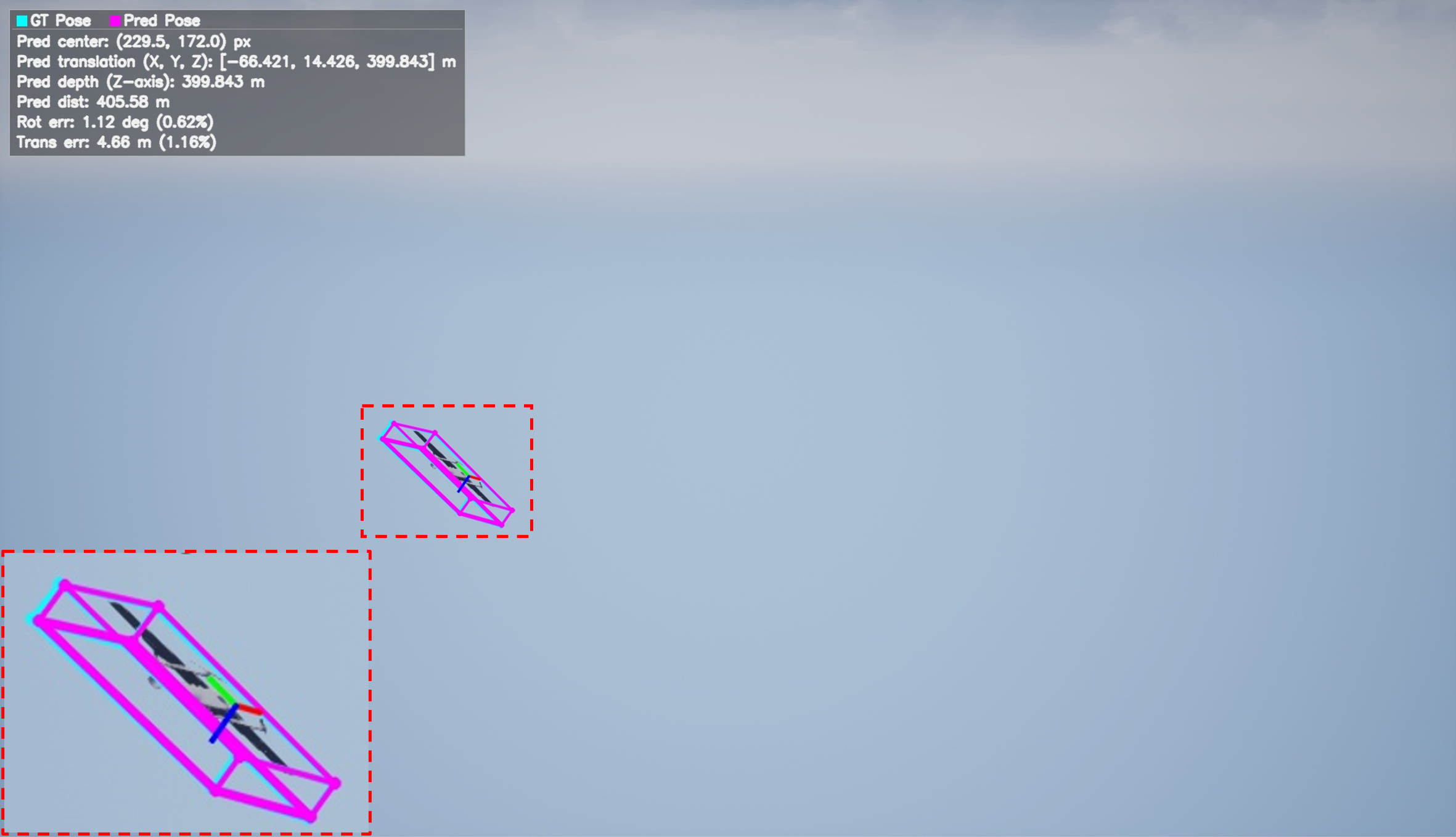}
\end{subfigure}
\hfill
\begin{subfigure}[t]{0.24\textwidth}
    \centering
    \includegraphics[width=\linewidth]{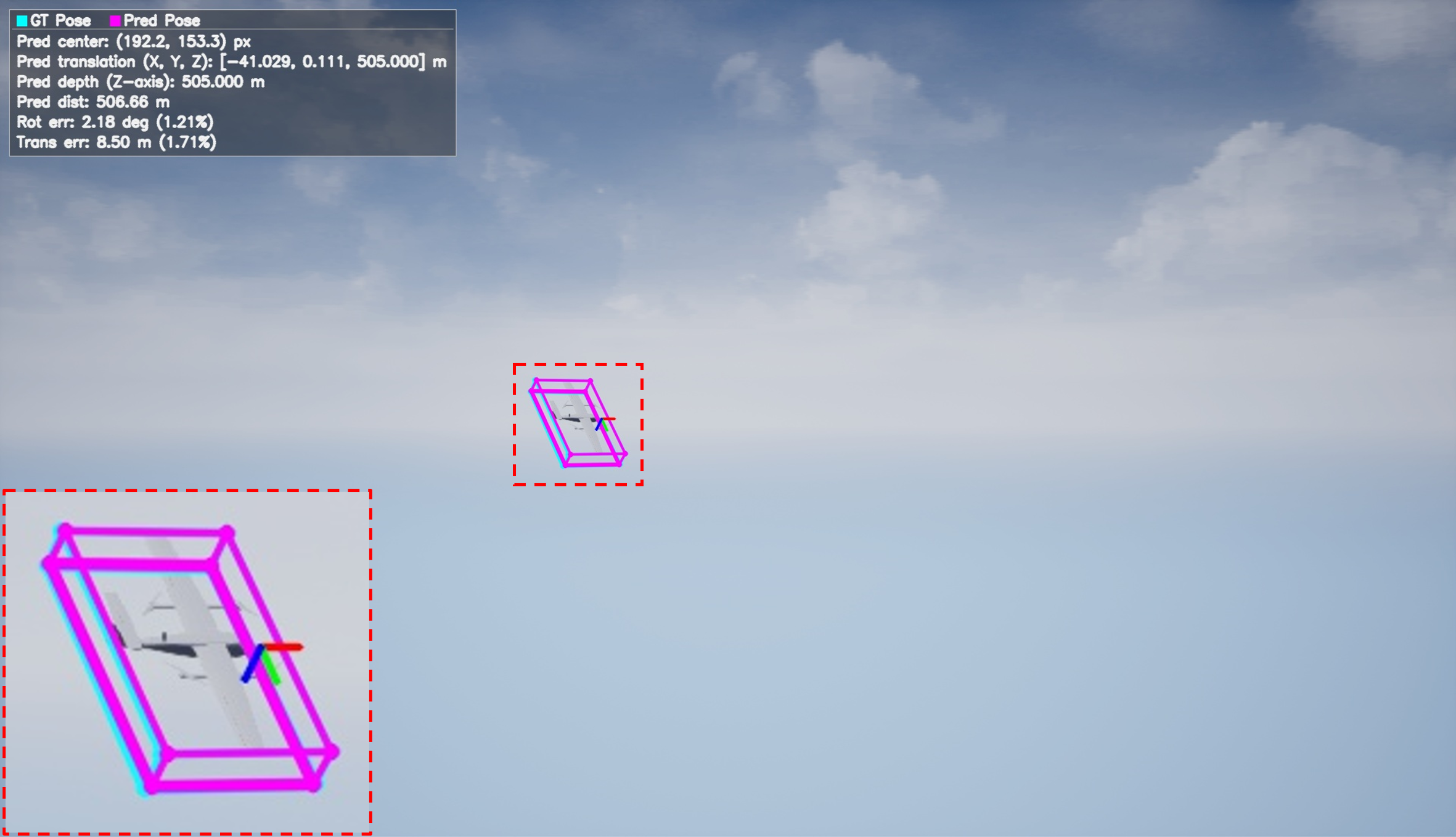}
\end{subfigure}
\hfill
\begin{subfigure}[t]{0.24\textwidth}
    \centering
    \includegraphics[width=\linewidth]{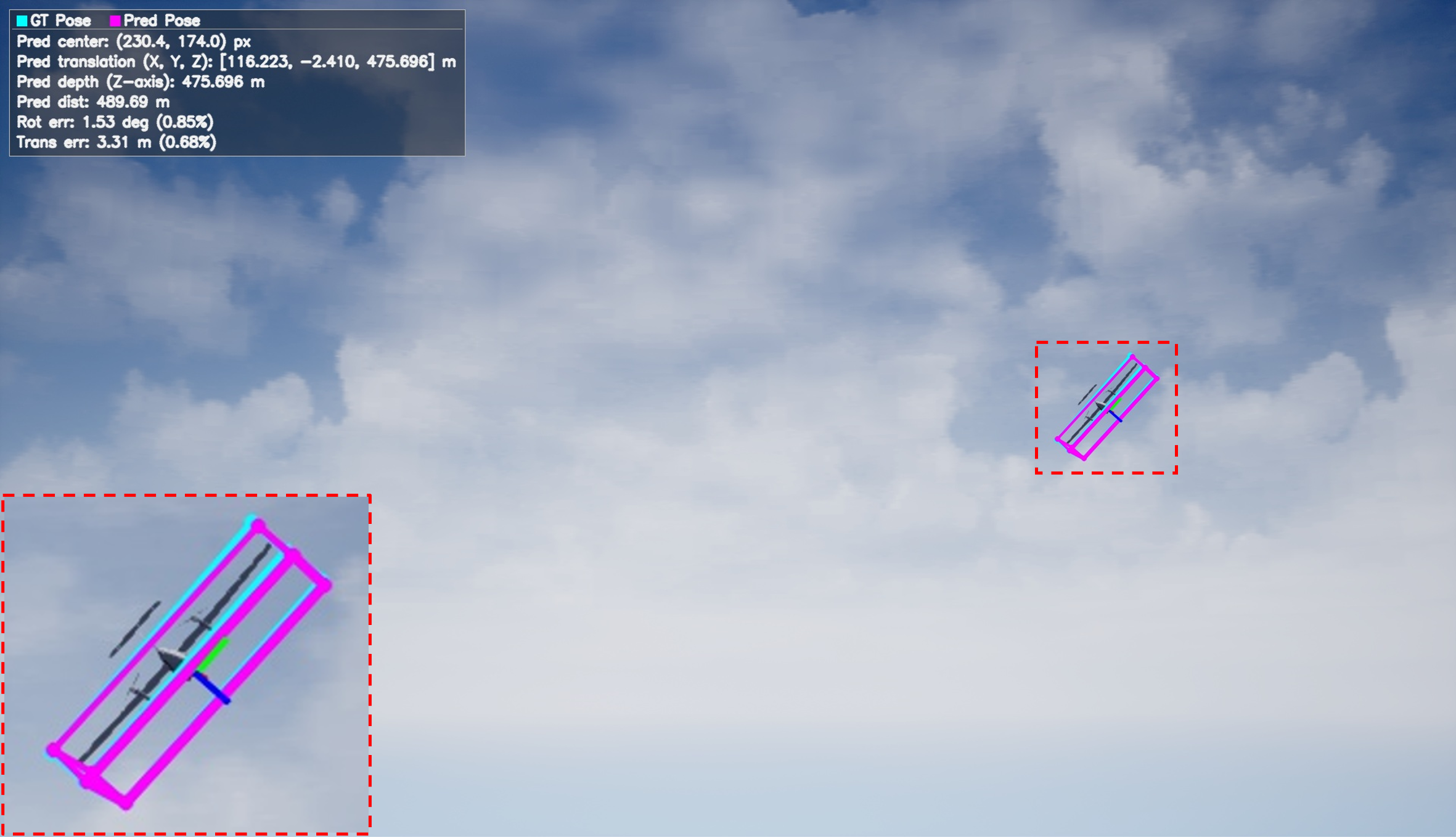}
\end{subfigure}
\hfill
\begin{subfigure}[t]{0.24\textwidth}
    \centering
    \includegraphics[width=\linewidth]{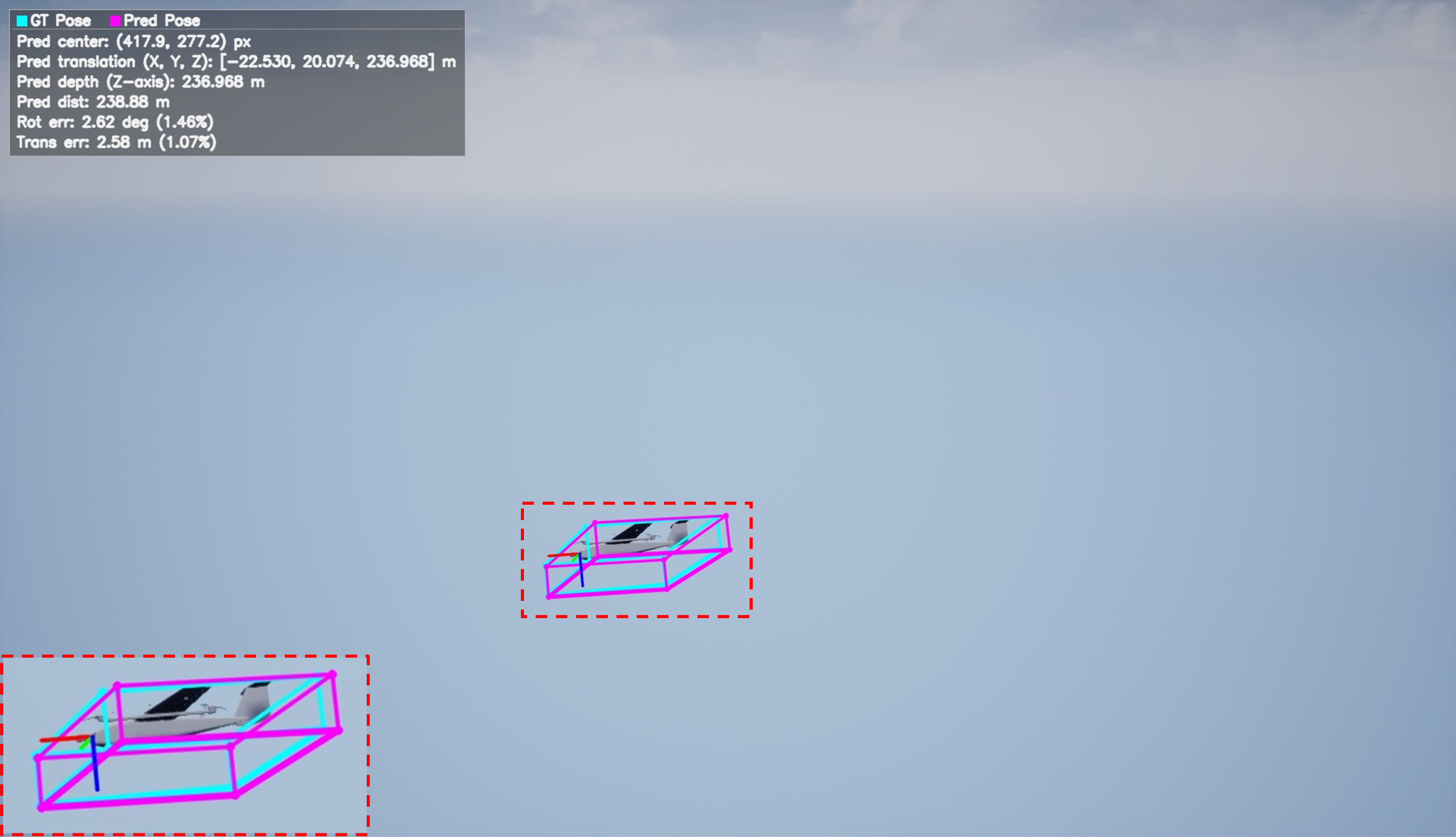}
\end{subfigure}
\caption{Qualitative visualization of 6-DoF pose estimation results produced by MF-UAVPose6D on the FW-UAV6DPose dataset. The cyan and magenta 3D bounding boxes denote the ground-truth and predicted poses, respectively.}
\label{Fig4}
\end{figure*}
To comprehensively evaluate the performance of MF-UAVPose6D in long-range 6-DoF pose estimation for fixed-wing UAVs, this paper compares it with multiple representative methods on the FW-UAV6DPose test set, covering strategies based on CAD models, reference images, detection assistance, and model-free settings. All results are obtained using official open-source implementations.

As shown in Table \ref{taball}, without requiring CAD models, depth maps, reference images, or external 2D detection boxes, MF-UAVPose6D achieves a mean rotation error of 4.96°, a mean translation error of 7.28 m, and a Pose@10/5\% of 82.8\%, attaining the best performance in terms of rotation accuracy and joint pose success rate. Compared with DronePose and GDR-Net, which rely on CAD models, the proposed method effectively reduces pose estimation errors while significantly reducing dependence on prior information, demonstrating good adaptability to long-range non-cooperative UAV observation scenarios.

Among methods without CAD model constraints, MF-UAVPose6D also shows clear advantages. Compared with Gen6D, which relies on reference data, and PoET, which relies on 2D detection boxes, the proposed method achieves consistent improvements in the main accuracy metrics. Compared with SC6D-Pose, MF-UAVPose6D reduces the mean rotation error from 9.36° to 5.15° and improves Pose@10/5\% from 73.7\% to 82.8\%, indicating stronger robustness and discriminative capability in orientation modeling and joint pose evaluation. Although SC6D-Pose achieves better results in mean translation error and ADD-0.5d, it relies on additional 2D detection box inputs; in contrast, MF-UAVPose6D requires only an RGB image and camera intrinsics to perform end-to-end pose estimation. Meanwhile, MF-UAVPose6D achieves a single-frame inference time of 4.6ms, only slightly higher than the 2.3ms of DronePose and significantly faster than GDR-Net, PoET, and SC6D-Pose. In summary, MF-UAVPose6D achieves a better balance among estimation accuracy, prior dependence, and inference efficiency, and can better meet the practical requirements of monocular 6-DoF pose estimation for long-range fixed-wing UAVs.

\subsection{Ablation Analysis of Key Components}

The performance of the proposed method mainly depends on three key factors: 1) whether the heatmap-guided target center localization module can provide a stable target anchor, 2) whether observation-ray modeling in PAM can effectively introduce viewpoint-aware geometric constraints, 3) whether DTS can complement local topological structure cues of long-range fixed-wing targets, and 4) whether the APDD in the translation branch can improve depth estimation accuracy. Based on these considerations, we analyze the contribution of each core component to overall 6-DoF pose estimation performance from three perspectives: center heatmap visualization, removal of observation-ray modeling,  ablation of dynamic topological sampling, and replacement of APDD with direct depth regression. 

\subsubsection{Reliability Analysis of the Target Center Anchor}

The target center anchor $(u,v)$ is the starting point of the subsequent geometric reasoning in MF-UAVPose6D, and its reliability directly determines the input quality of PAM, DTS, and translation back-projection. To evaluate whether this anchor remains stable under long-range conditions, we visualize the target-center heatmaps produced by the heatmap-guided target center localization module. For fixed-wing UAVs, the target usually occupies only a small image region, while the sky background often exhibits weak texture, high brightness, and illumination variations, which may interfere with the center response. If the peak of the target center heatmap is unstable, errors will propagate downstream to subsequent critical processes. Therefore, the reliability of the target center anchor is crucial to the entire pipeline.

\begin{figure}[!t]
	\centering
	\includegraphics[width=\linewidth]{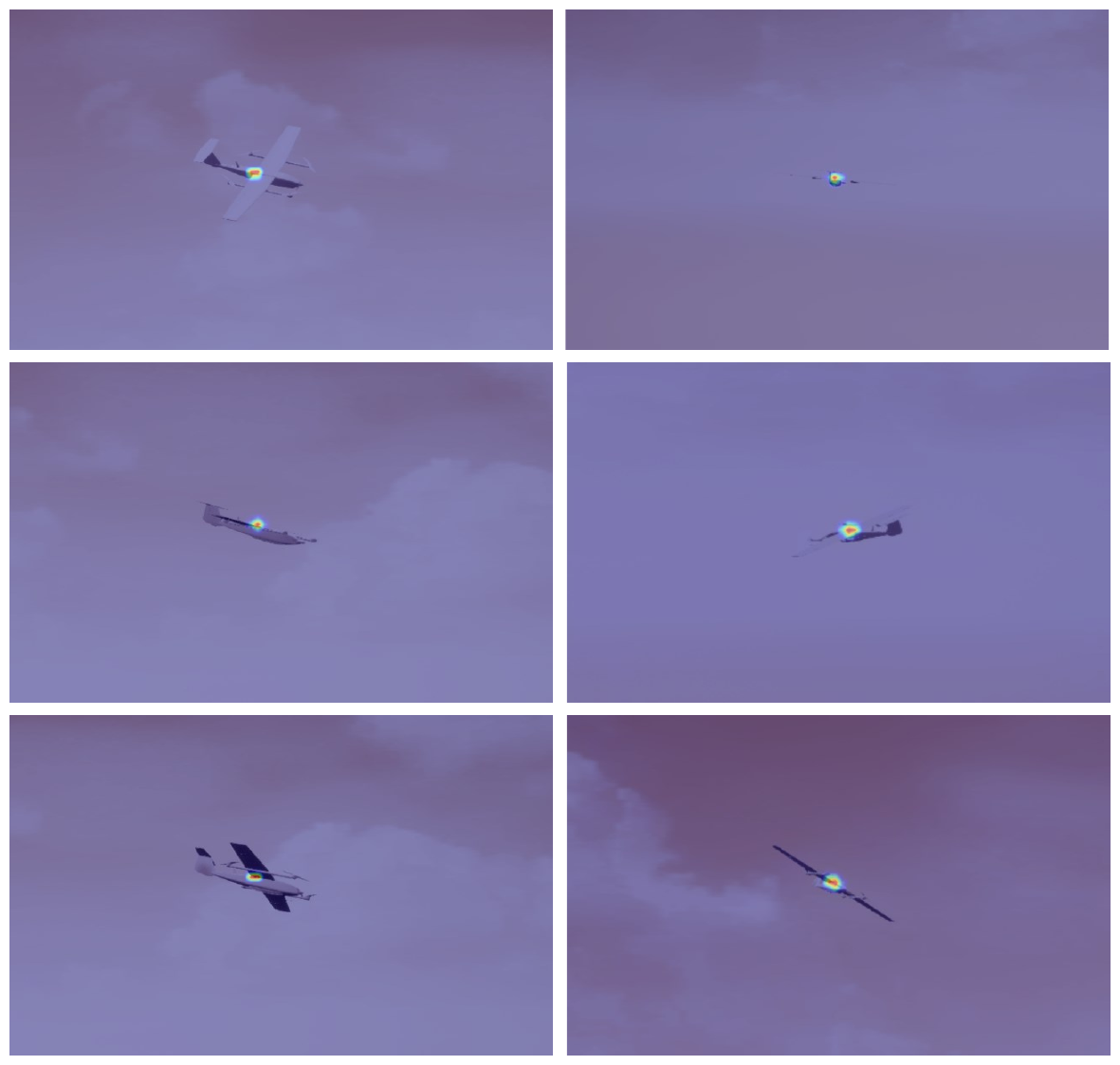}	
	\caption{Visualization of heatmap-guided object center localization. The high-response regions are accurately concentrated around the UAV target center, while background responses are effectively suppressed.}
	\label{Fig_heatmap}	
\end{figure}

As shown in Fig. \ref{Fig_heatmap}, the high-response regions predicted by the model are consistently concentrated around the UAV body center while effectively suppressing background responses. Even when the target scale is small and the wing and tail edges are ambiguous, the peak of the target-center heatmap still falls near the center of the visible target region. This indicates that the network learns a center prior associated with the overall structure of fixed-wing UAVs, rather than relying on local textures or background brightness responses. These results demonstrate that heatmap-guided target-center localization can provide a stable target-center anchor under long-range weak-texture backgrounds, offering reliable input for subsequent geometry-aware pose decoding.

\subsubsection{Ablation Study on PAM}

This experiment aims to evaluate the contribution of PAM to long-range UAV rotation estimation. Specifically, the model with PAM injects the observation direction into the visual embedding through the unit observation ray $\mathbf{d}$, and further uses $\mathbf{d}$ in rotation decoding to construct $R_{\mathrm{ray}}$, which transforms $R_{\mathrm{allo}}$ into $R_{\mathrm{ego}}$ in the camera coordinate system. The ablated model without PAM removes the above ray-related modeling. Except for this modification, all other settings of the two models are kept identical to objectively quantify the performance gain brought by PAM. The results are shown in Table \ref{tabpam} and Fig. \ref{figpam}.

\begin{table}[!t]
\centering
\caption{Effect of the PAM on 6-DoF Pose Estimation.}
\small
\setlength{\tabcolsep}{3pt}
\resizebox{\linewidth}{!}{
\begin{tabular}{l|c|c|c|c|c}
\hline
Methods & $E_r$ & Rot$<10^\circ$ & $E_t$ & Pose@10°/5\% & ADD-0.5d \\
\hline
\makecell[l]{MF-UAVPose6D\\without PAM} & 6.91° & 82.4\% & 7.03m & 75.8\% & 89.5\% \\
\hline
\makecell[l]{MF-UAVPose6D\\with PAM}    & 4.96° & 90.9\% & 7.28m & 82.8\% & 88.5\% \\
\hline
\end{tabular}
}
\label{tabpam}
\end{table}

\begin{figure}[!t]
	\centering
	\includegraphics[width=\linewidth]{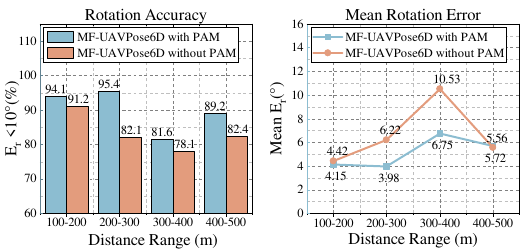}	
	\caption{The left subplot shows the proportion of samples with $E_r<10^\circ$ across distance ranges, while the right subplot presents the corresponding mean rotation error.}
	\label{figpam}	
\end{figure}

The results in Table \ref{tabpam} show that PAM substantially improves rotation estimation accuracy and further enhances overall pose determination. Compared with the model without PAM, the PAM-enabled model reduces the mean rotation error from 6.91° to 4.96°, and improves Rot$<10^\circ$ from 82.4\% to 90.9\%. Meanwhile, Pose@10°/5\% increases from 75.8\% to 82.8\%. These results indicate that PAM effectively incorporates viewing-angle information into object representation and rotation decoding via $\mathbf{d}$, enabling the model to better distinguish perspective variations caused by changes in viewing-angle from intrinsic object pose changes.

The distance-wise results in Fig. \ref{figpam} further demonstrate that the benefits of PAM are mainly observed in mid- to long-range scenarios. In the 200-300m, 300-400m, and 400-500m ranges, introducing PAM improves Rot$<10^\circ$ by 7.7\%, 0.5\%, and 6.7\%, respectively, while reducing the mean rotation error by 1.44° and 3.42° in the 200-300m and 300-400m ranges, respectively. This suggests that the gain brought by PAM becomes more pronounced as the target distance increases. The reason is that, under long-range conditions, the target scale decreases and structural cues from the wings, fuselage, and tail gradually weaken. As a result, purely visual features are more likely to misinterpret perspective variations induced by viewing-angle changes as intrinsic target rotations. PAM explicitly introduces $\mathbf{d}$ and provides the rotation branch with a observation-direction prior related to imaging geometry, thereby reducing the impact of viewing-angle variation on pose regression. Although the ablated model achieves a slightly higher Rot$<10^\circ$ in the 100-200m range, its mean rotation error remains larger than that of the full model, indicating that PAM still yields more stable estimation under continuous rotation-error metrics. Overall, PAM provides more robust pose constraints when appearance structures become less discriminative and viewpoint ambiguity increases, thereby effectively improving the robustness of mid- to long-range UAV rotation estimation.

\subsubsection{Ablation Study on DTS}
The DTS ablation experiment is conducted to evaluate the role of the eight-point topological context in fixed-wing UAV pose estimation. In the full model, edge midpoints and corner points are sampled based on the predicted 2D scale, and the resulting topological context is used to guide dense regional representation. In contrast, the ablated model removes this context pathway and directly feeds the scale-driven dense regional features into the translation and rotation decoders. All other settings are kept identical, and the distance-wise comparison results are shown in Fig. 4.

\begin{figure}[!t]
	\centering
	\includegraphics[width=\linewidth]{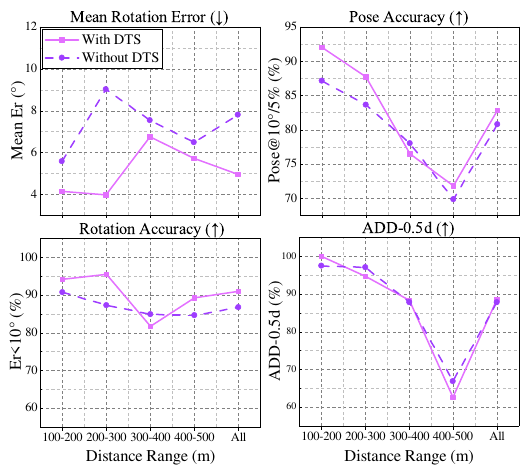}	
	\caption{The ablation results of DTS across different range intervals. DTS improves rotation estimation and overall pose accuracy, validating the effectiveness of topological context for long-range pose estimation.}
	\label{figpam}	
\end{figure}

As shown in the figure, DTS brings a notable improvement in rotation estimation. Overall, introducing DTS reduces the mean rotation error from 7.79° to 4.96°, increases Rot<10° from 86.7\% to 90.9\%, and improves Pose@10°/5\% from 80.8\% to 82.8\%. This indicates that dense regional features can capture the overall appearance of the target, but are insufficient to fully model the key structural relationships of fixed-wing UAVs under long-range imaging conditions. The distance-wise results further show that DTS consistently improves rotation accuracy across different ranges. In particular, at 100–200 m and 200–300 m, the mean rotation error is reduced by 1.43° and 5.03°, respectively, while Pose@10°/5\% is improved by 4.9 and 4.1 percentage points. At the longer range of 400–500 m, Rot<10° also increases from 84.6\% to 89.2\%. In contrast, the improvement in ADD-0.5d is less consistent, suggesting that DTS does not primarily contribute to translation estimation. Instead, it injects structured topological cues, such as wing contours, fuselage outlines, and tail layouts, into the regional visual representation $f_{\mathrm{reg}}$, thereby enhancing pose robustness for long-range small targets.

\subsubsection{Ablation Study of APDD}
This ablation study evaluates the effect of APDD on long-range depth recovery and translation estimation. The experiment only changes the depth estimation strategy: Direct-Z directly regresses absolute-depth, whereas APDD predicts an implicit physical scale and recovers object depth by combining it with focal length and 2D apparent scale. All other modules and experimental settings are kept identical to isolate the independent contribution of the perspective-scale constraint.

\begin{table}[!t]
\centering
\caption{Effect of APDD on Depth and Translation Estimation.}
\small
\setlength{\tabcolsep}{3pt}
\begin{tabular}{l|c|c|c|c|c}
\hline
Methods  & $E_t$  & Depth$<5\%$ & $E_z$ & Pose@10$^\circ$/5\%  & ADD-0.5d \\
\hline
Direct-Z  & 7.67m  & 87.2\% & 7.43m & 79.4\% & 87.2\% \\
\hline
APDD      & 7.28m  & 89.0\% & 6.96m  & 82.8\% & 88.5\% \\
\hline
\end{tabular}
\label{tabapdd}
\end{table}

\begin{figure}[!t]
	\centering
	\includegraphics[width=\linewidth]{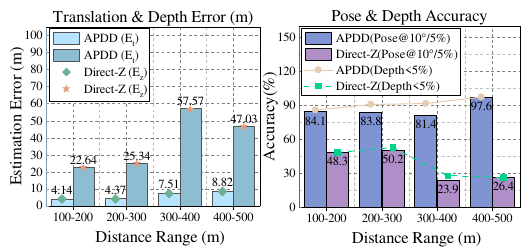}	
	\caption{Performance comparison between APDD and Direct-Z across different target distance ranges. The left subplot shows the translation and depth errors in each distance range, and the right subplot presents the corresponding proportions of samples satisfying Depth$<5\%$ and Pose@10°/5\%.}
	\label{figapdd}	
\end{figure}

As shown in Table \ref{tabapdd}, APDD substantially improves depth estimation, translation estimation, and overall pose estimation performance. Compared with Direct-Z, APDD reduces the mean translation error from 7.76m to 7.28m and the depth error ($E_z$) from 7.43m to 6.96m, while increasing Depth<5\% from 87.2\% to 89.0\%. Meanwhile, Pose@10°/5\% and ADD-0.5d improve from 79.4\% and 87.2\% to 82.8\% and 88.5\%, respectively. These results indicate that APDD significantly improves 3D translation estimation through more stable depth recovery, thereby increasing the success rate under strict pose evaluation criteria.

The distance-wise results further demonstrate the effectiveness of APDD in long-range scenarios. As shown in Fig. \ref{figapdd}, Direct-Z suffers from evident performance degradation in the 400–500m range, where the translation error increases to 16.58 m and Pose@10°/5\% drops to only 62.6\%. In contrast, APDD maintains a lower depth error in this range, with an $E_z$ of 15.59 m, and achieves an ADD-0.5d of 71.8\%, corresponding to a 4.8\% improvement over Direct-Z. In addition, APDD also improves ADD-0.5d by 0.7\% in the 300–400m range. These findings demonstrate that APDD effectively mitigates the instability of direct absolute-depth regression at long distances by incorporating perspective-scale constraints among focal length, apparent scale, and implicit physical scale. It is therefore a key module for improving long-range UAV translation estimation and strict pose estimation success rates.

\subsection{Sequential 6-DoF Pose Estimation Analysis}

Previous experiments validate the pose estimation performance of MF-UAVPose6D on single-frame images. To further evaluate its temporal consistency under continuous observation, this paper conducts sequence-level analysis on multiple continuous flight sequences from the test set and selects two representative sequences to visualize the frame-wise variations of the predicted 6-DoF pose as well as the 3D translation trajectories. Here, the root mean square error (RMSE) denotes the error between the predicted and ground-truth values of the corresponding translation or rotation component across all frames in the sequence, and is used to quantify the trend consistency of the continuous pose trajectory.

\begin{figure}[!t]
	\centering
	\includegraphics[width=\linewidth]{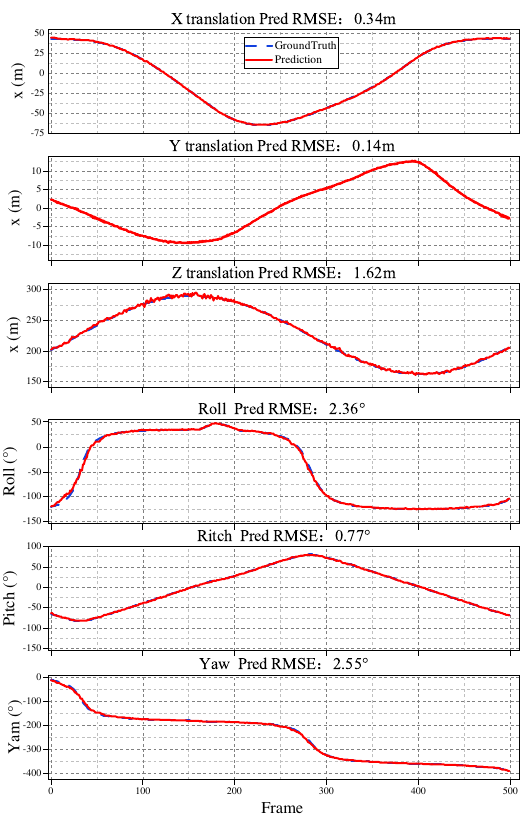}	
	\caption{6-DoF pose estimation results for the first sequence case. The temporal variations of the 3D translation components $(X, Y, Z)$ and rotation angles $(Roll, Pitch, Yaw)$ are presented across the frame sequence.}
	\label{figseq1}	
\end{figure}

\begin{figure}[!t]
	\centering
	\includegraphics[width=\linewidth]{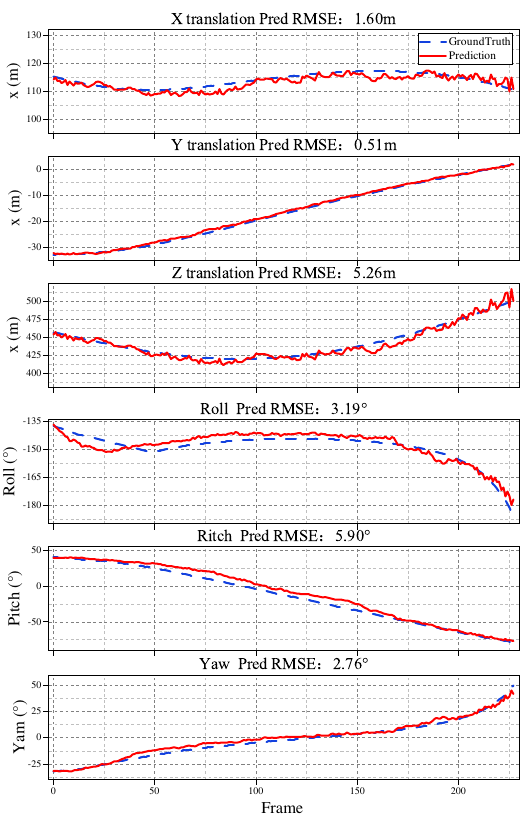}	
	\caption{6-DoF pose estimation results for the first sequence case.}
	\label{figseq2}	
\end{figure}

\begin{figure}[!t]
\centering
\begin{subfigure}[t]{0.48\linewidth}
    \centering
    \includegraphics[width=\linewidth]{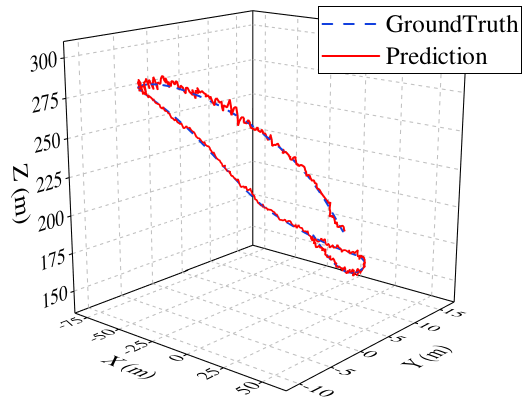}
\end{subfigure}
\hfill
\begin{subfigure}[t]{0.48\linewidth}
    \centering
    \includegraphics[width=\linewidth]{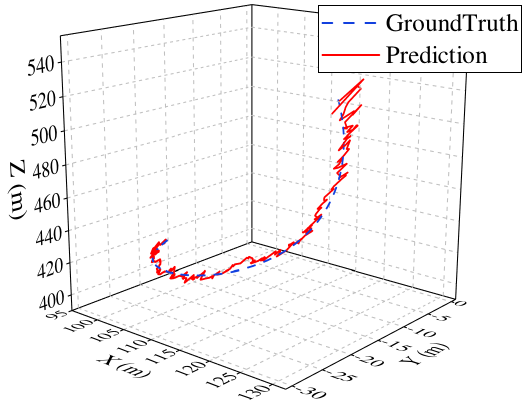}
\end{subfigure}
\caption{3D translation trajectory estimation results for the first and second sequence cases. The blue dashed lines denote the ground-truth 3D trajectories, while the red solid lines represent the predicted trajectories.}
\label{fig3d}
\end{figure}

As shown in Fig. \ref{figseq1}, Sequence 1 covers an observation range of 100–300m, and its predicted results maintain good consistency with the ground truth in terms of the overall variation trend. The RMSE values of the translation components X, Y, and Z are 0.34 m, 0.14 m, and 1.62 m, respectively, while those of the rotation components Roll, Pitch, and Yaw are 2.36°, 0.77°, and 2.55°, respectively. These results indicate that MF-UAVPose6D can accurately recover the target pose under short- to medium-range conditions and stably track pose variations across consecutive frames. Fig. \ref{figseq2} further presents the estimation results of Sequence 2 under the long-range condition of 400–500m. Due to the reduced target scale and weakened texture information, the translation RMSE increases to 1.60 m, 0.51 m, and 5.26 m, while the rotation RMSE values are 3.19°, 5.90°, and 2.76°, respectively, showing a certain degree of long-range degradation. Nevertheless, the predicted curves still preserve the variation trend of the ground-truth poses well. In addition, the 3D trajectory results in Fig. \ref{fig3d} show a high degree of overlap between the predicted and ground-truth trajectories in both sequences. Overall, MF-UAVPose6D can effectively maintain the consistency of pose variation trends under continuous observation conditions, providing further validation for its application in consecutive-frame UAV observation tasks.

\section{Conclusion}

This paper addresses non-cooperative observation scenarios of long-range fixed-wing UAVs and proposes MF-UAVPose6D, a model-free monocular 6-DoF pose estimation method. Taking heatmap-guided target center localization as the starting point of geometric reasoning, the proposed method explicitly constructs the unit observation ray through the target center via PAM, and introduces the observation-direction prior into target representation and rotation decoding to alleviate rotation ambiguity caused by viewing-angle variations in long-range targets. DTS is further employed to complement edge, corner, and local contour information of fixed-wing UAVs, while the proposed APDD reduces the instability of direct absolute-depth regression in long-range scenarios. Extensive experimental results demonstrate that the proposed method achieves favorable performance in rotation accuracy, depth recovery, and 3D translation estimation, while maintaining high inference efficiency, validating its effectiveness and practical potential for long-range monocular UAV 6-DoF pose estimation. Future research will extend to longer-range observation scenarios to further enhance the engineering applicability of model-free monocular 6-DoF pose estimation.

\bibliographystyle{IEEEtran}
\bibliography{references}

\end{document}